\documentclass[10pt,twocolumn,letterpaper]{article}

\usepackage{iccv}
\usepackage{times}
\usepackage{epsfig}
\usepackage{graphicx}
\usepackage{amsmath}
\usepackage{amssymb}

\usepackage{enumitem}
\usepackage{lipsum}
\usepackage{url}
\usepackage{booktabs}
\usepackage{caption}
\usepackage{color}
\usepackage{subcaption}
\DeclareMathAlphabet{\mathbfit}{OML}{cmm}{b}{it}
\usepackage{tabularx}
\newcolumntype{C}{>{\centering\arraybackslash}X}
\usepackage[export]{adjustbox}
\usepackage{float}
\usepackage{nameref}
\usepackage{hangg}

\newcommand{\model}{\text{DPG}}

\newcommand{\bfsection}[1]{{\noindent \textbf{#1} \quad}}

\usepackage[
    pagebackref=true,
    breaklinks=true,
    letterpaper=true,
    colorlinks=true,
    bookmarks=false
    ]{hyperref}

\iccvfinalcopy 

\def\iccvPaperID{69}

\ificcvfinal\fi
\begin{document}

\title{Disentangling Propagation and Generation for Video Prediction}
\author{
    Hang Gao\thanks{Equal contribution.}$_1$
    \quad
    Huazhe Xu\footnotemark[1]$_1$
    \quad
    Qi{-}Zhi Cai\thanks{Work was done while Q.C. and R.W. were at Berkeley
    DeepDrive.}$_2$
    \quad
    Ruth Wang\footnotemark[2]$_3$
    \quad
    Fisher Yu$_1$
    \quad
    Trevor Darrell$_1$ \\
    UC Berkeley$_1$
    \quad
    Sinovation Ventures AI Institute$_2$
    \quad
    Columbia University$_3$
    }

\maketitle

\begin{abstract}
A dynamic scene has two types of elements: those that move fluidly and can be
predicted from previous frames, and those which are disoccluded (exposed) and
cannot be extrapolated. Prior approaches to video prediction typically learn
either to warp or to hallucinate future pixels, but not both. In this paper, we
describe a computational model for high-fidelity video prediction which
disentangles motion-specific propagation from motion-agnostic generation. We
introduce a  confidence-aware warping operator which gates the output of pixel
predictions from a flow predictor for non-occluded regions and from a context
encoder for occluded regions. Moreover, in contrast to prior works where
confidence is jointly learned with flow and appearance using a single network,
we compute confidence after a warping step, and employ a separate network to
inpaint exposed regions. Empirical results on both synthetic and real datasets
show that our disentangling approach provides better occlusion maps and produces
both sharper and more realistic predictions compared to strong baselines.
\end{abstract}

\section{Introduction}
\begin{figure}
    \begin{center}
        \includegraphics[width=\linewidth]{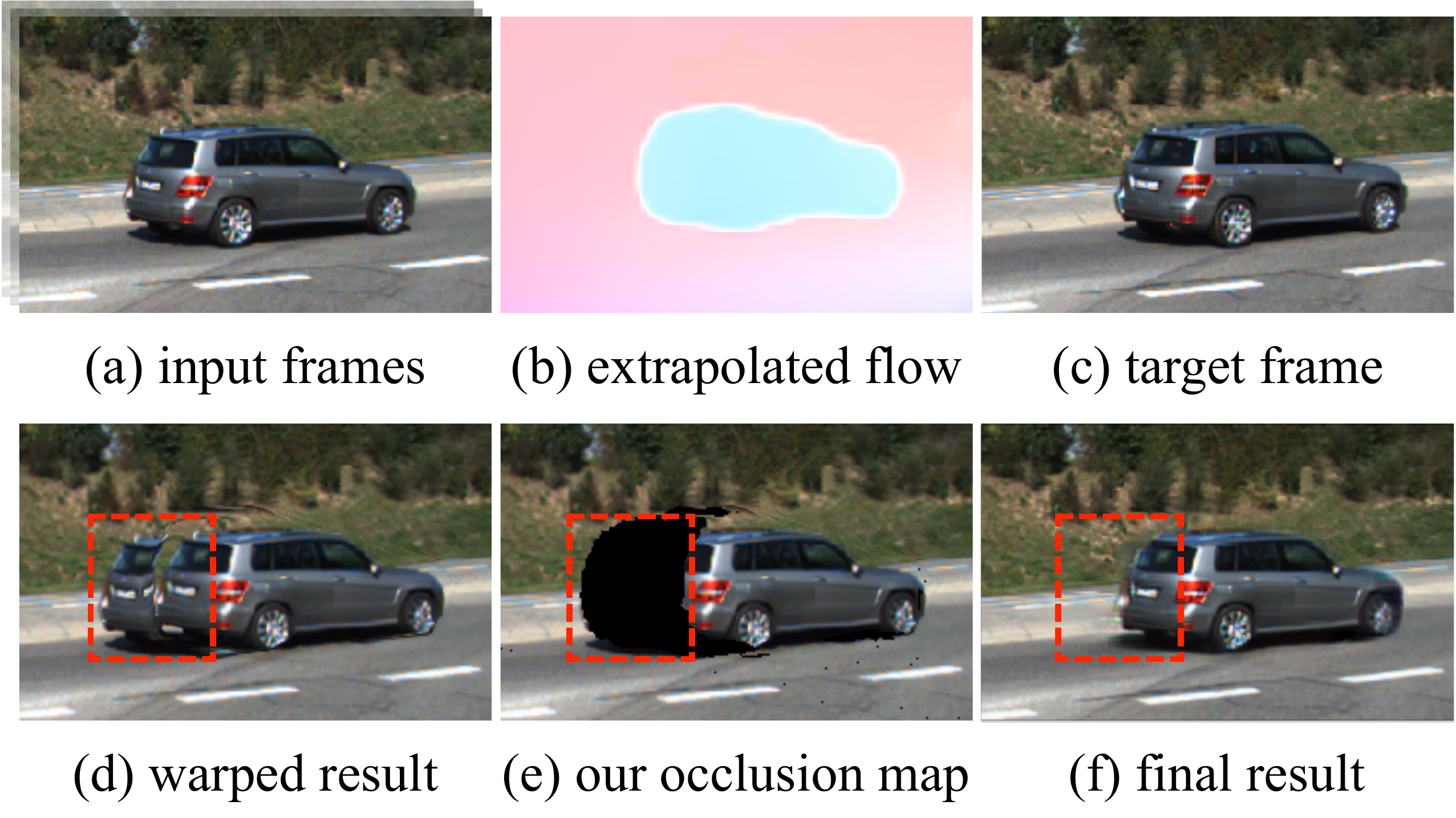}
    \end{center}
    \caption{Video prediction by extrapolating the motion and hallucinating the
    pixels to be exposed. Given a video whose last frame shown in \textbf{(a)},
    our model first extrapolates optic flow \textbf{(b)} to its near future
    target \textbf{(c)}. Direct motion propagation introduces ghosting effects
    \textbf{(d)} on pixels to be disoccluded (marked with the bounding box). Our
    disentangling approach computes a sparse occlusion map \textbf{(e)}  and
    uses an contextual encoder to inpaint low-confidence patches, producing the
    final prediction \textbf{(f)}.}
    \label{fig:teaser}
\end{figure}

\begin{figure*}[th!]
    \begin{center}
    \includegraphics[width=0.95\textwidth]{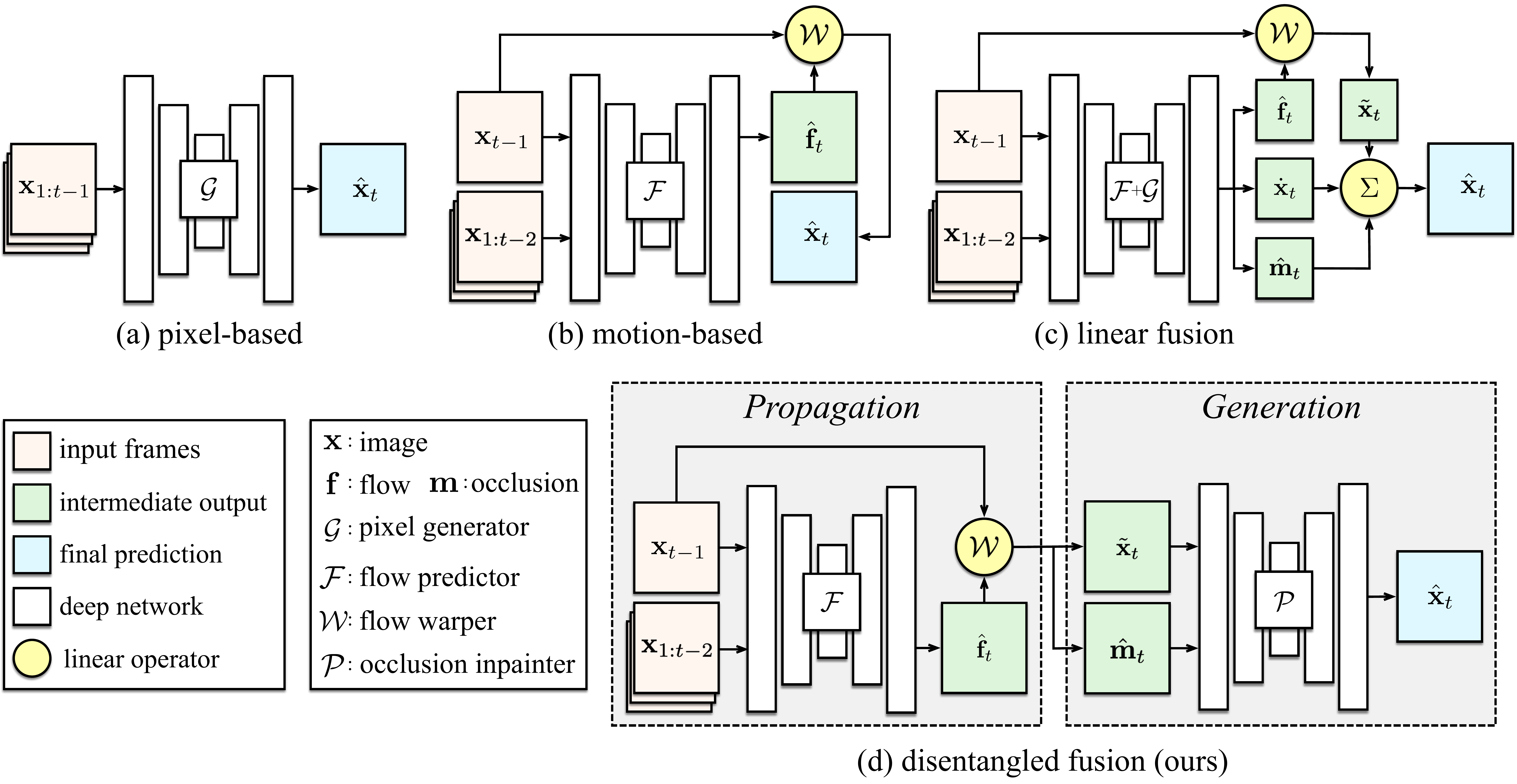}
    \end{center}
    \caption{Overview of previous works and our proposed frameworks. Given
    $(t-1)$ frames and to predict one frame, previous video prediction methods
    typically consist of a flow predictor $\cF$, or a pixel generator $\cG$, or
    both: \textbf{(a)} \textit{pixel-based} approaches;
    \textbf{(b)} \textit{motion-based} approaches; \textbf{(c)}
    \textit{linear-fusion} models; \textbf{(d)} ours.}
    \label{fig:overview}
\end{figure*}

Video prediction is a challenge due to the many varied factors that combine to
generate future appearance.

State-of-the-art approaches to video prediction are often purely
 \textit{pixel-based} and  generate each pixel from scratch
 (Figure~\hyperref[fig:overview]{2a}) \cite{lotter2016deep,
 villegas2017learning,tulyakov2017mocogan,villegas2017decomposing,
 denton2018stochastic,wichers2018hierarchical,byeon2017contextvp}. They rely on
 3D or recurrent convolutional networks to encode spatial contents over time,
 hoping to capture implicit motion representations. Another active line of
 purely \textit{motion-based} approaches seeks pixel correspondence for explicit
 motion propagation \cite{pintea2014deja,liu2017video}; in this approach
 standard networks are augmented with an optical-flow representation and a
 warping function (Figure~\hyperref[fig:overview]{2b}). Because all pixels are
 copy-pasted from the history image buffer, a degree of temporal consistency and
 spatial richness are automatically ensured. However, disocclusions---locations
 in the target frame to which extrapolated optical flow has no
 projection---cause severe errors with this family of models (Figure
 \hyperref[fig:teaser]{1e}).

We observe that humans can easily predict  future appearance (e.g., in a scene
such as shown in  Figure~\ref{fig:teaser}) by tracking an object's past motion
and extrapolating future pixel values, as well as  imagine the appearance of
disoccluded pixels (the background) based on prior knowledge of similar scenes
previously observed. We propose a compositional approach that disentangles
spatial and temporal prediction, using a post-warp occlusion map to mediate
between flow-based prediction and inpainting-based reconstruction of disoccluded
regions.

%
%
%



Disentangling is different from the \textit{linear fusion}
 approaches~\cite{finn2016unsupervised,hao2018controllable} that use a single
 multi-task network to learn \textit{pixel-} and \textit{motion-based}
 predictions at the same time. To compose their results, the networks may also
 learn an extra occlusion map for weighted summation over each pixel
 (Figure~\hyperref[fig:overview]{2c}). Though those approaches are effective in
 many cases, it is unclear how much pixel and flow predictions can benefit each
 other. As our experiments indicate on the real dataset, these two tasks do not
 actually correlate significantly with each other.


We propose a compositional framework that explicitly disentangles
motion-specific propagation and motion-agnostic generation into two submodules,
and thus called \model{}, for high-quality video prediction. We factorize flow
and pixel predictions into a serialized pipelines. We first employ a flow
predictor for extrapolating optical flows and then an occlusion inpainter for
context encoding based on images warped by flows. After flow prediction and
warping, we compute confidence directly based on predicted flow rather than via
joint unsupervised learning. The downstream occlusion inpainter generates pixels
on occluded areas based on confidence map.

We evaluate our approach on both standard CalTech Pedestrian
dataset~\cite{dollar2012pedestrian} and more challenging KITTI Flow
dataset~\cite{menze2015object} with larger motions and occlusions. Our approach
achieves new state-of-the-art performance on both datasets with large
performance gain on the perceptual realism metrics. To quantitatively evaluate
our computed occlusion confidence maps, we perform ablation study on the
RoamingImages~\cite{Janai2018ECCV} dataset where occlusion ground-truth masks
are available. Our model again compare favorably against previous baselines in
occlusion prediction in terms of Intersection over Union (IoU) metrics.

In summary, our main contributions are in two aspects: a fusion pipeline that
utilizes both optical flow and image synthesis for video prediction; a post-warp
confidence-based occlusion map computation to disentangle pixel propagation and
generation.

\section{Related Works}

\subsection{Photo-realistic Image Synthesis}
Realism is the constant pursuit of high-quality image
synthesis~\cite{johnson2016perceptual,chen2014semantic,dosovitskiy2016generating,wang2004image,li2016combining,xie2016top}.
Recent developments towards photo-realistic image synthesis typically feature
Generative Adversarial Networks (GANs) \cite{goodfellow2014generative}.
Conditioned on categorical labels~\cite{brock2018large}, textual
descriptions~\cite{reed2016generative} or
segmentations~\cite{wang2018pix2pixHD}, high-fidelity images are shown to be
able to synthesized. The closest work to ours are Dense Pose
Transfer~\cite{neverova2018dense} and Transformation-grounded View
Synthesis~\cite{park2017transformation}, which generally generate images by
warping the original image with a learned appearance flow, conditioned on either
given view transformation angle or a predefined dense pose, and then inpaint the
ambiguous parts. In our case, however, the model needs to \textit{predict}
future motion and further synthesize based on both spatial and temporal
information. 

\noindent \textbf{Spatial Context Encoding} \quad Pixels are not isolated. On
the contrary, there are many
cases~\cite{bertalmio2000image,barnes2009patchmatch,zhang2018context} where
appropriate spatial contexts must be retrieved. Spatial context
encoders~\cite{pathak2016context} query learned dataset priors with exposed
appearance in search for missing patches. To our task, the ``mask'' is an
occlusion map where motion predictions are erroneous. Particularly, we employ
partial convolutions \cite{liu2018image} in our occlusion inpainter's encoding
blocks. Different from prior works on static images, our approach leverages not
only the spatial but also temporal context.

\subsection{High-fidelity Video Prediction}
Different from image synthesis, video prediction not only cares for per-frame
visual quality but also cross-frame consistency. Recent
approaches~\cite{denton2018stochastic,byeon2017contextvp,lotter2016deep,ranzato2014video,
saito2017temporal} are typically \textit{pixel-based}, which generate each pixel
from scratch using implicit motion representations. Such methods may suffer from
blurry effects, especially in the presence of unseen novel scenes. On the
contrary, \textit{motion-based} methods~\cite{pintea2014deja,liu2017video} excel
in making sharp predictions, yet fail in occlusion areas where motion
predictions are erroneous or ill-defined. Meanwhile, Reda \textit{et
al.}~\cite{reda2018sdc} propose to model moving appearances with both
convolutional kernels as in \cite{finn2016unsupervised} and vectors as optical
flow. Our closest prior work is \cite{hao2018controllable} which also composes
the pixel- and flow-based predictions through occlusion maps. However, our
proposed method can differentiate in three aspects: (1) our pixel and flow
prediction tasks are separately trained; (2) we employ an occlusion inpainter
for our pixel generation so that more contextual information after warping can
be utilized; (3) instead of predicting occlusion as another side task, we
directly refer to predicted flows as the proxy for post warping confidence. It
should be noted that \cite{wang2018occlusion} also seek for occlusion analytics
to further parse flow reliability.

\noindent \textbf{Disentangling Motion and Content} \quad Since videos are
essentially moving contents whose semantics are agnostic to motions -- one
person's identity does not change while he's walking -- it is natural to
disentangle them apart.
\cite{tulyakov2017mocogan,villegas2017decomposing,denton2018stochastic} are the
three representative works in such direction. Though sharing the intuition to
disentangle motion with other video property, our approach builds on lower-level
vision such as pixel correspondence. We take advantage of optical flow and their
embeded occlusion clues to naturally factorize pixel and flow predictions into
separate modules.

\section{Approach}



\begin{figure*}[th!]
    \begin{center}
    \begin{subfigure}[t]{0.35\linewidth}
        \includegraphics[width=\linewidth]{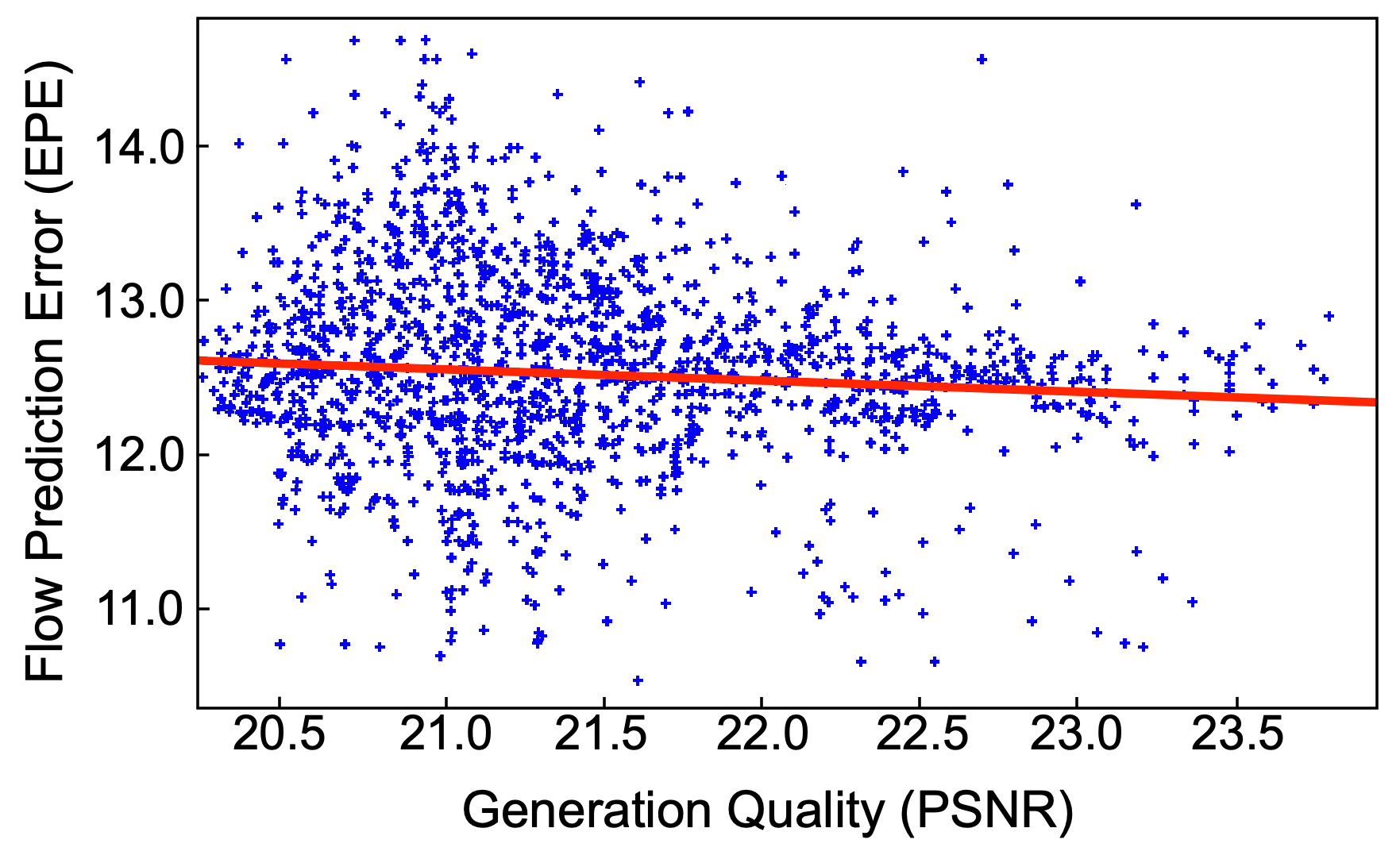}
        \caption{linear fusion~\cite{hao2018controllable}}
        \label{fig:corr}
    \end{subfigure}
    \quad
    \quad
    \begin{subfigure}[t]{0.35\linewidth}
        \includegraphics[width=\linewidth]{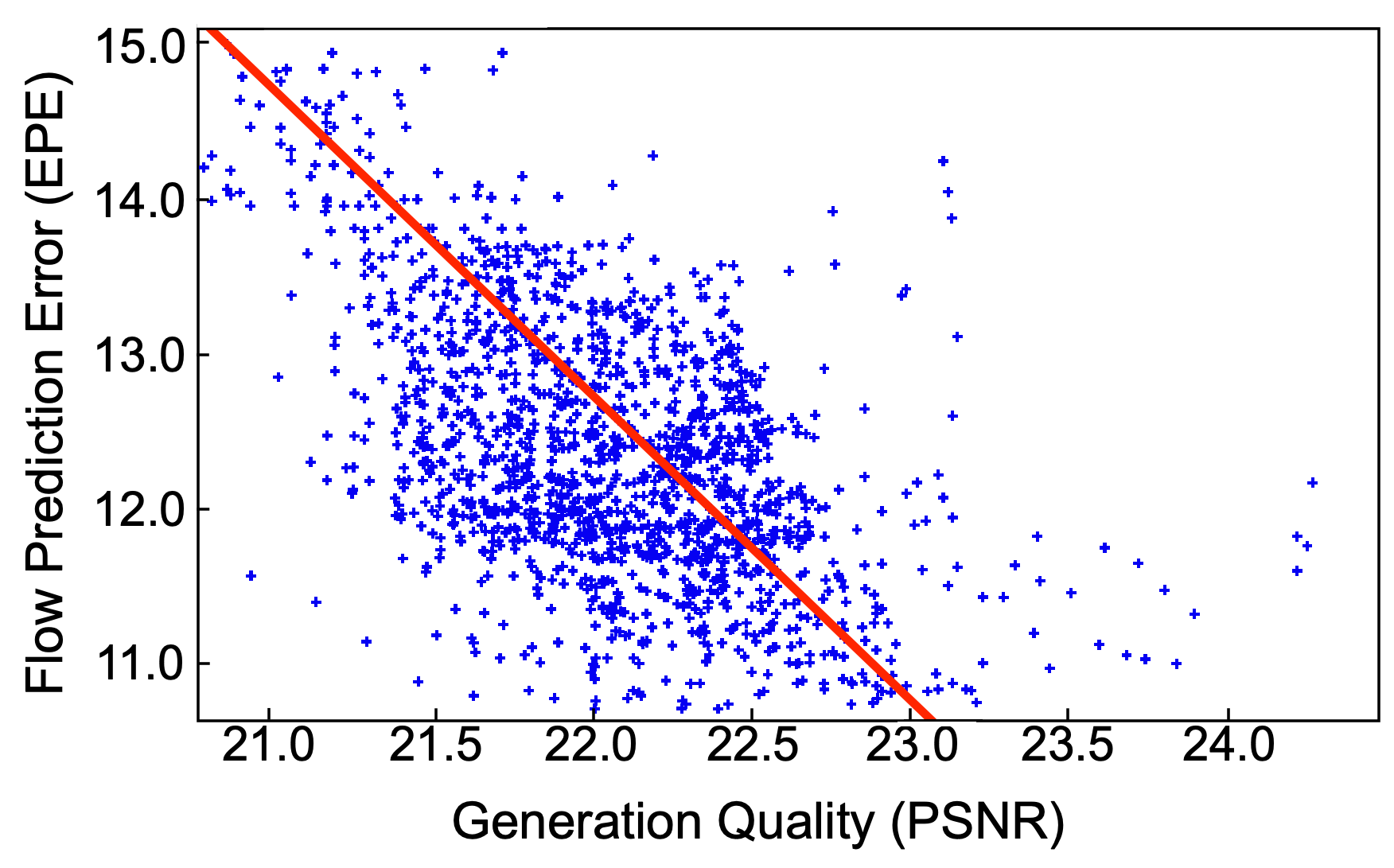}
        \caption{disentangled fusion (ours)}
        \label{fig:corr-out}
    \end{subfigure}
    \end{center}
    \vspace{-0.5cm}
    \caption{The correlation between pixel and flow prediction tasks evaluated by different models. We show scatter plots about generation quality versus flow prediction error on KITTI Flow dataset~\cite{menze2015joint}. Pixel and flow prediction tasks do not actually correlate significantly with each other on previous models using multi-task learning. Our disentangled fusion pipeline shows that after factorizing these tasks apart, both tasks could be better learned.}
\end{figure*}

\label{sec:overview}

Video prediction aims to synthesize future frames given a stack of history frames. For the ease of exposition, we here focus on a
$\left(t-1\right)$-in $1$-out prediction task: given an input video sequence denoted as $\vx_{1:t-1}$, the model aims to predict the frame $\vx_{t}$  which should be accurate and visually sharp. In this paper, we build a model beyond simply using $\ell_1$ or $\ell_2$ reconstruction loss, which is widely known to encourage blurry results. Concretely, as illustrated in Figure \hyperref[fig:overview]{2d}, our model is disentangled into two orthogonal yet complementary modules -- a flow predictor $\cF$ and an occlusion inpainter $\cP$ to learn flow and pixel predictions separately.

Given the history frames $\vx_{1:t-1}$, the flow predictor $\cF$ learns to predict the flow field $\hat{\vf}_t$ for the pixel correspondence between the last input frame $\vx_{t-1}$ and the target $\vx_{t}$. Utilizing $\hat{\vf}_t$, we can propagating the input frame $\vx_{t-1}$ into a motion-dependent prediction $\tilde{\vx}_{t}$. Since it neglects the existence of occlusions, we additionally compute an occlusion map $\hat{\vm}_t$ during the propagation. Based on $\tilde{\vx}_{t}$ and $\hat{\vm}_t$, the inpainting module $\cP$ learns to inpaint occluded region from a learned prior within a training set. By explicitly disentangling future dynamics to motion-dependent areas and motion-agnostic areas, our model is able to predict high-fidelity future frames. 

\subsection{Motion Propagation} \label{sec:fp}

Our module $\cF$ computes the correlation of appearances between each pair of frames from the history inputs and predicts future motion dynamics using optical flow. We choose it over other motion representations, such as frame differences \cite{huang2018makes} or sparse trajectories \cite{hao2018controllable} because it provides rich information about motion occlusions over pixels.

\begin{figure}[t]
    \begin{center}
    \begin{subfigure}[b]{0.46\linewidth}
        \includegraphics[width=\linewidth]{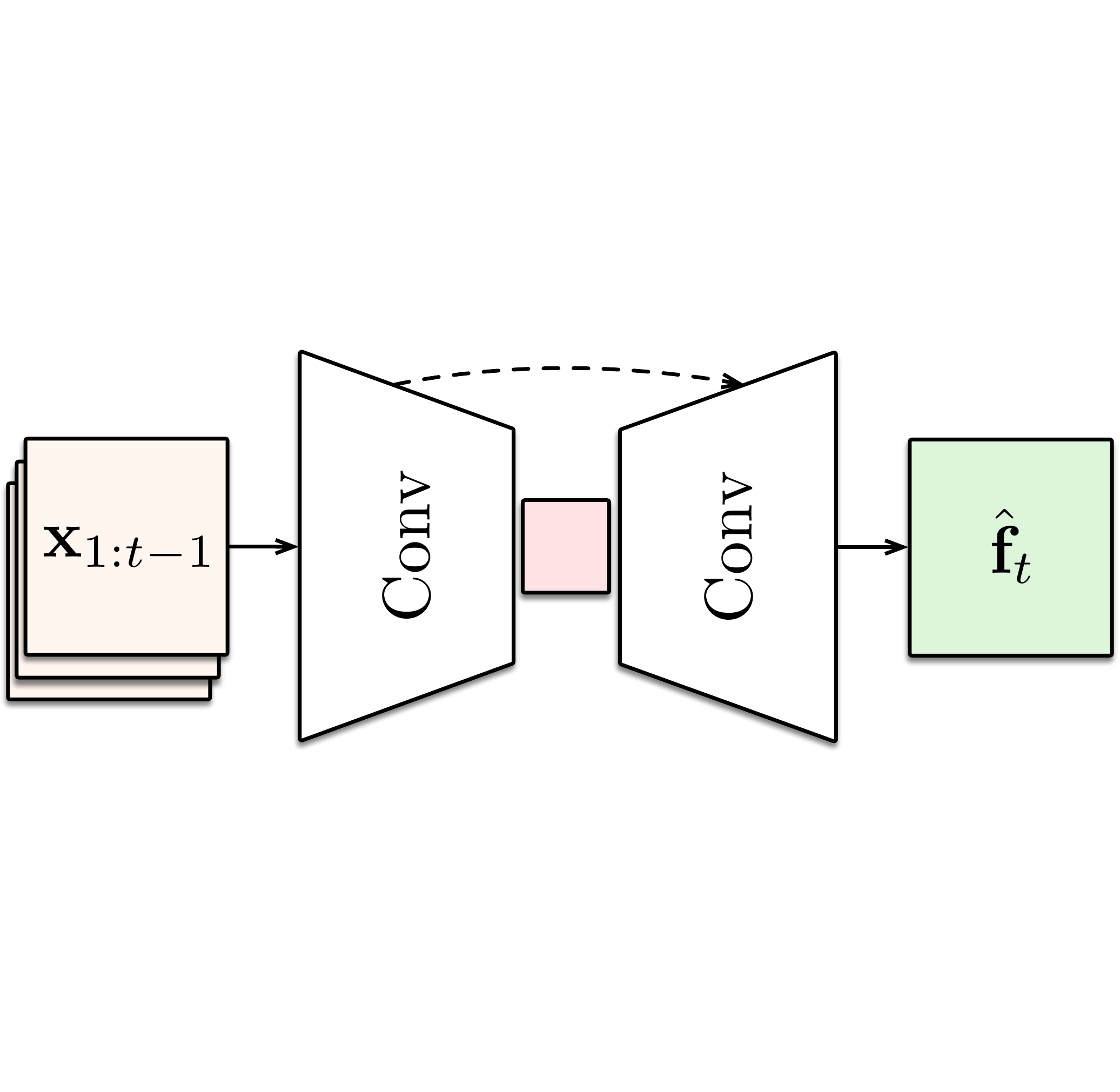}
        \caption{Flow predictor $\cF$.}
        \label{fig:ff1}
    \end{subfigure} \,
    \begin{subfigure}[b]{0.46\linewidth}
        \includegraphics[width=\linewidth]{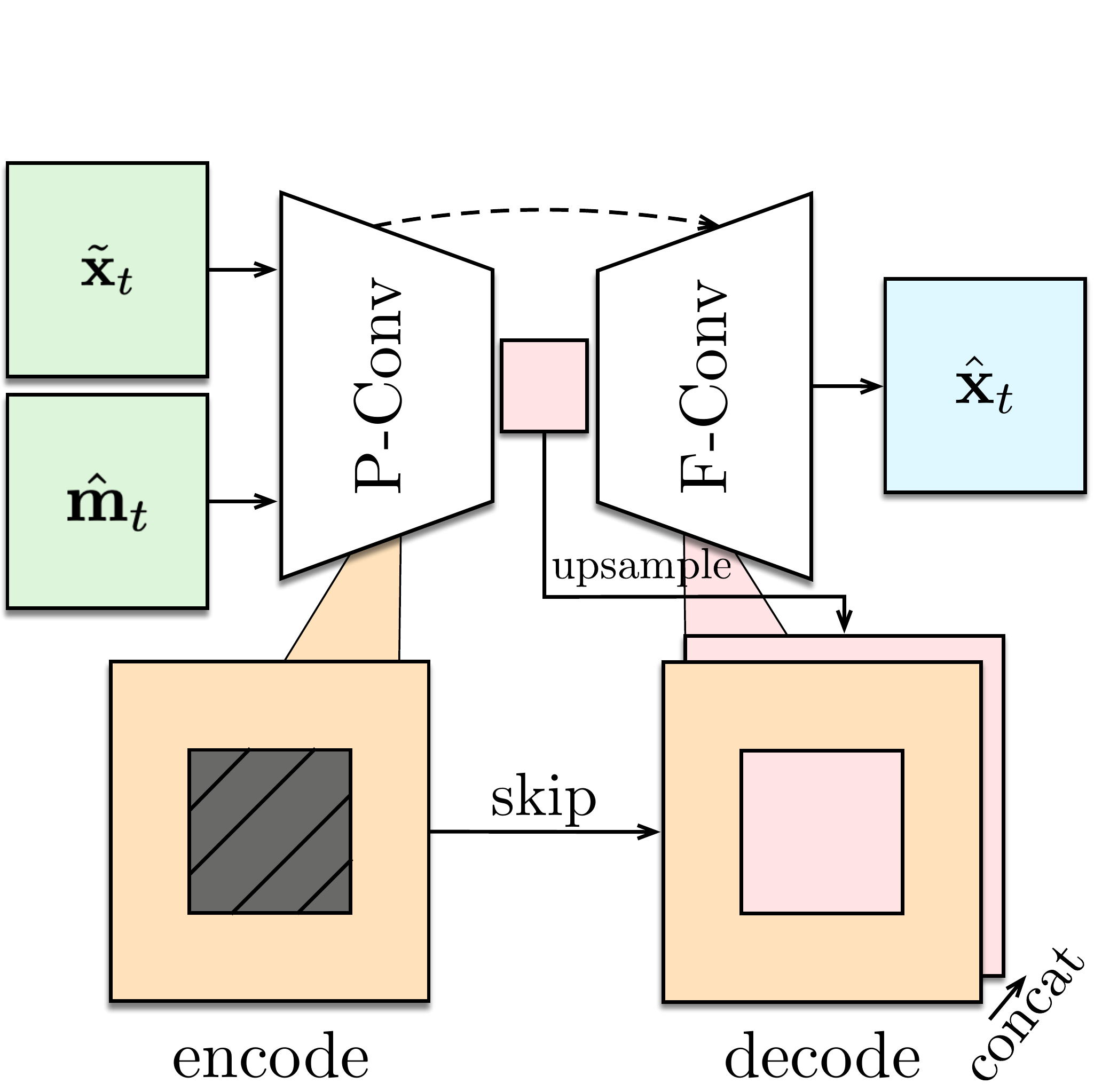}
        \caption{Occlusion inpainter $\cP$.}
        \label{fig:ff2}
    \end{subfigure}
    \\
    \end{center}
    \caption{Conceptual illustrations of our $\cF$ and $\cP$. \textbf{(a)} Our propagation module $\cF$ is a standard encoder-decoder fully convolutional network with skip connections, which takes in history frames $\vx_{1:t-1}$ and predict the backward flow $\hat{\vf}_t$ from $\vx_t$ to $\vx_{t-1}$. \textbf{(b)} taking in the computed occlusion map $\hat{\vm}_t$ and the warped image $\tilde{\vx}_t$, our inpainting module $\cP$ inpaints for the final prediction $\hat{\vx}_t$. It replaces standard convolutional blocks with partial convolutions (P-Conv) and fusion blocks (F-Conv) in its encoder and decoder, respectively.}
    \label{fig:arch}
\end{figure}

As illustrated in Figure \hyperref[fig:arch]{3a}, the module $\cF$ is an encoder-decoder network with skip connections. The output of $\cF$ is a 2-channel flow field $\hat{\vf}_t$ that aims to propagate the last frame $\vx_{t-1}$ into the predicted target frame $\tilde{\vx}_{t}$. Formally, let $\{(i, j)\} \in \vx_{t}$ be a Cartesian grid over the target frame, and we have

\begin{equation}
    \hat{\vf}_t = \{(\Delta_i, \Delta_j)\} = \cF(\vx_{1:t-1}).
\end{equation}

By assuming local linearity, we can 
sample the future frame from the last given frame as
\begin{equation}
    \tilde{\vx}_t =
    \cS(\vx_{t-1}; \hat{\vf}_t),
\end{equation}
where $\cS(\cdot; \hat{\vf}_t)$ is a bilinear sampler that generates the new image by first mapping the regular grid to the transformed grid and then bilinearly interpolating between produced sub-pixels.

For training the module $\cF$, we adopt a masked pixel loss $\cL_p$ and a smooth loss $\cL_{smt}$ which is similar to previous works on unsupervised flow estimation \cite{meister2017unflow,yin2018geonet}

\begin{equation}
    \label{recon_loss}
    \begin{aligned}
    \cL_{p}(\tilde{\vx}_{t}, \vx_{t}; \hat{\vm}_t) 
    &= 
    \alpha \frac{1 - SSIM(\tilde{\vx}_{t} \odot \hat{\vm}_t, \vx_{t} \odot \hat{\vm_t})}2 \\ 
    &+
    (1 - \alpha) \|\tilde{\vx}_{t} \odot \hat{\vm}_t - \vx_{t} \odot \hat{\vm}_t\|_1
    \end{aligned}
\end{equation}
\begin{equation}
    \begin{aligned}
    \cL_{smt}(\hat{\vf}_t, \vx_t) 
    &= 
    \sum_{i, j} |\nabla \hat{\vf}_t(i, j)|
    \cdot (e^{
    - |\nabla \vx_t(i, j)|
    })^T,
    \end{aligned}
\end{equation}
where $SSIM(\cdot, \cdot)$ denotes structural similarity index, $\odot$ denotes element-wise product, $\nabla$ is a vector differential operator, $T$ denotes the transpose of image gradient weighting, and $\alpha$ is our trade-off weight between the loss terms, which is fixed at $0.9$ through cross-validations.

The training loss for module $\cF$ is formulated as
\begin{align}
    \cL_{\cF} = \cL_{p}(\tilde{\vx}_{t}, \vx_{t}; \hat{\vm}_t)+\lambda_{smt} \cL_{smt}(\hat{\vf}_t, \vx_t),
\end{align}
where $\lambda_{smt}$ is a hyper-parameter that control the training schedule, we use $\lambda_{smt}=0.1$ through a coarse grid search.

\subsection{Occlusion Map}
\label{sec:map}
\begin{figure}[tp]
    \centering
    \begin{subfigure}[b]{0.4\linewidth}
        \includegraphics[width=\linewidth]{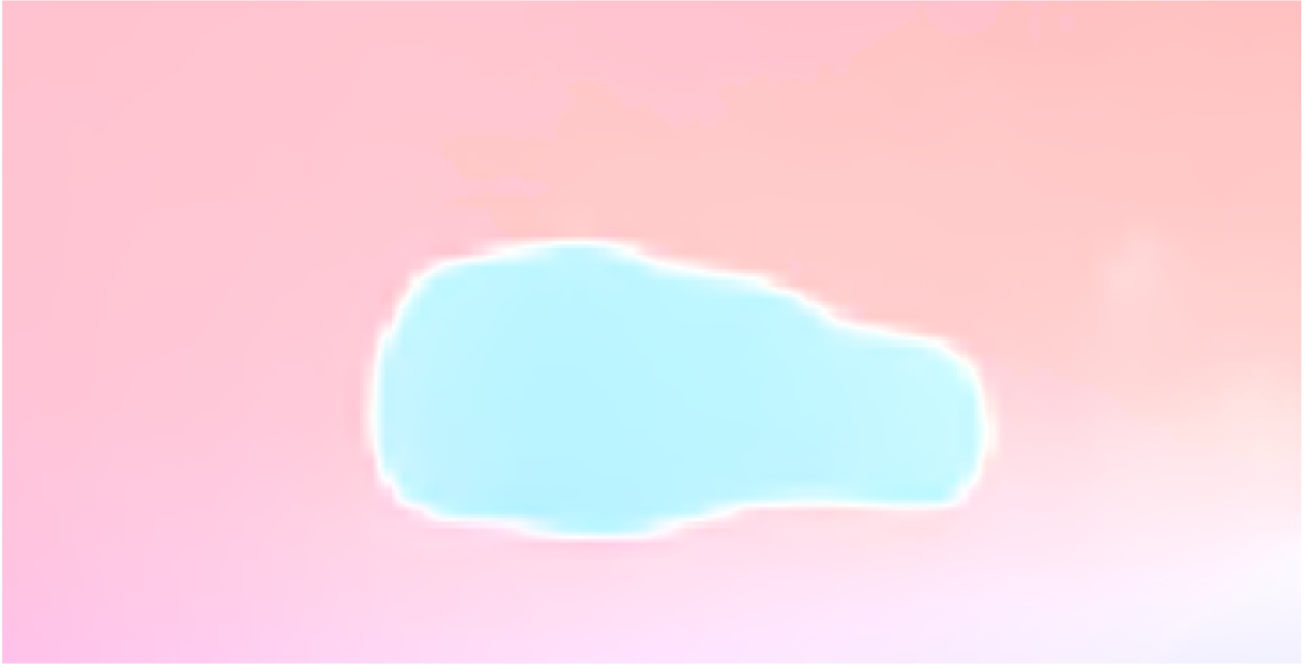}
        \caption{$t=0$, flow.}
        \label{fig:ff1}
    \end{subfigure} \quad
    \begin{subfigure}[b]{0.4\linewidth}
        \includegraphics[width=\linewidth]{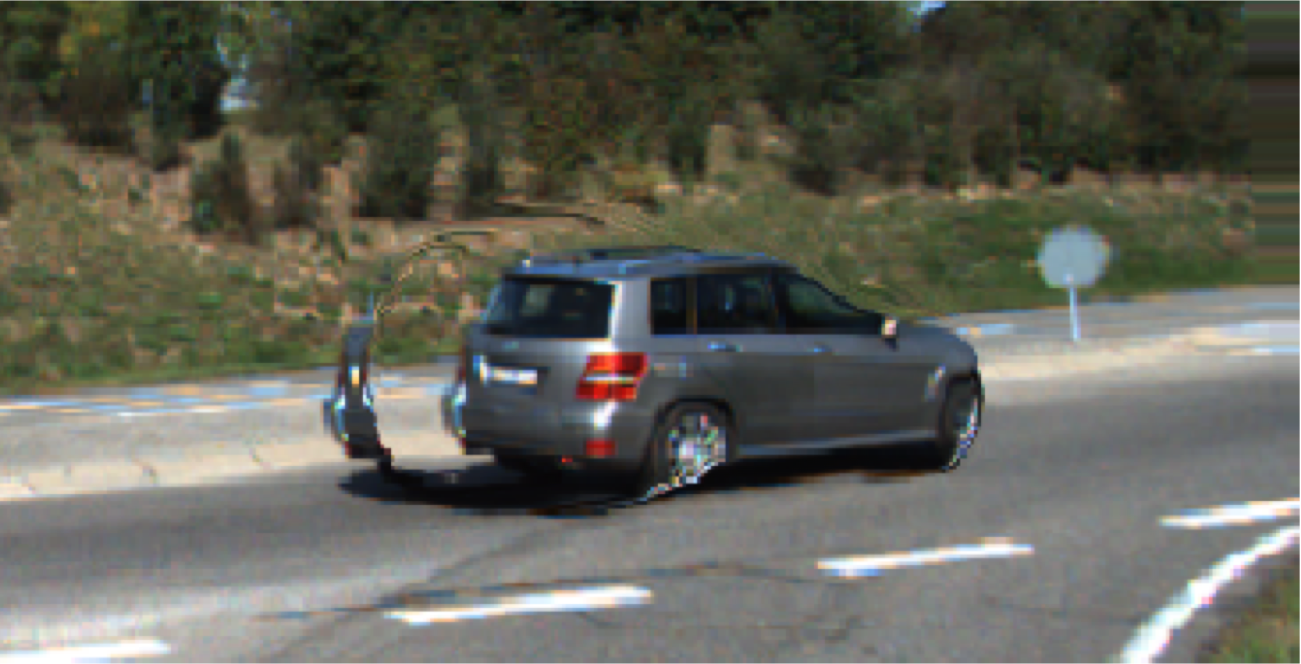}
        \caption{$t=1$, warping result}
        \label{fig:ff2}
    \end{subfigure}
    \\
    \begin{subfigure}[b]{0.4\linewidth}
        \includegraphics[width=\linewidth]{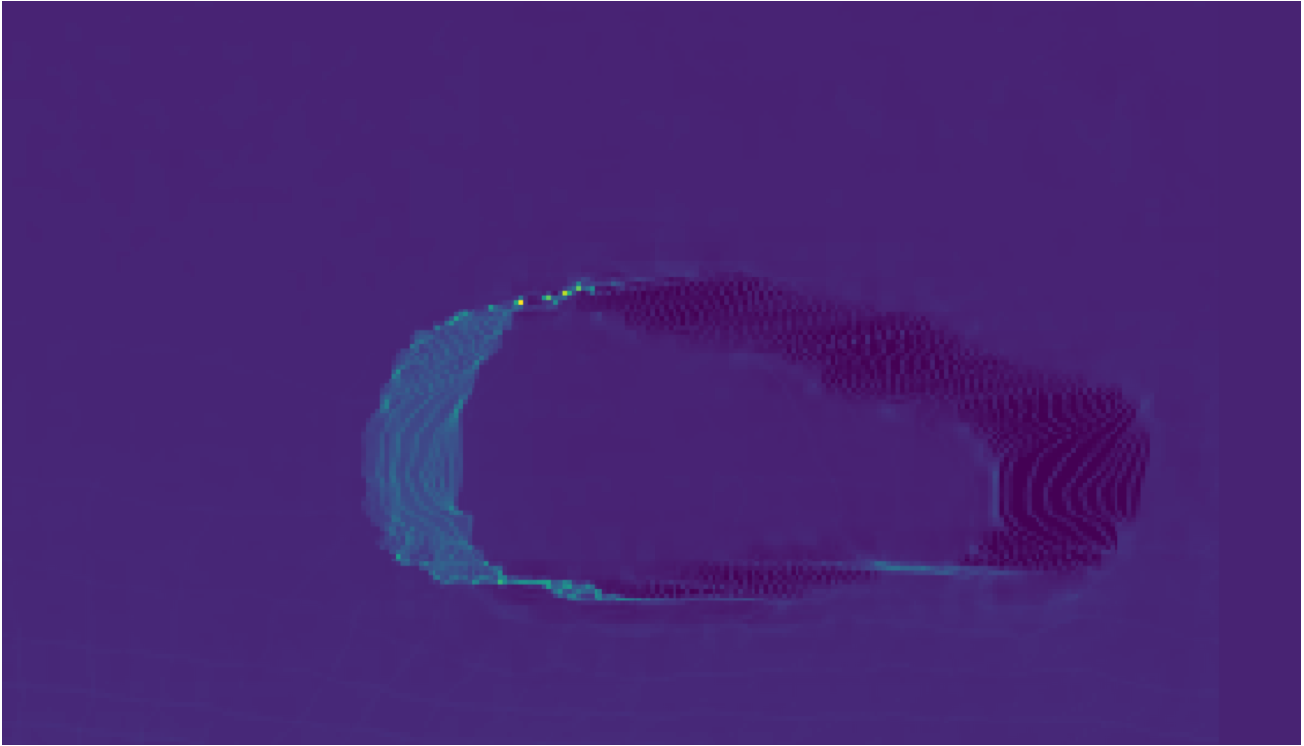}
        \caption{$t=1$, density map.}
        \label{fig:ff3}
    \end{subfigure} \quad
    \begin{subfigure}[b]{0.4\linewidth}
        \includegraphics[width=\linewidth]{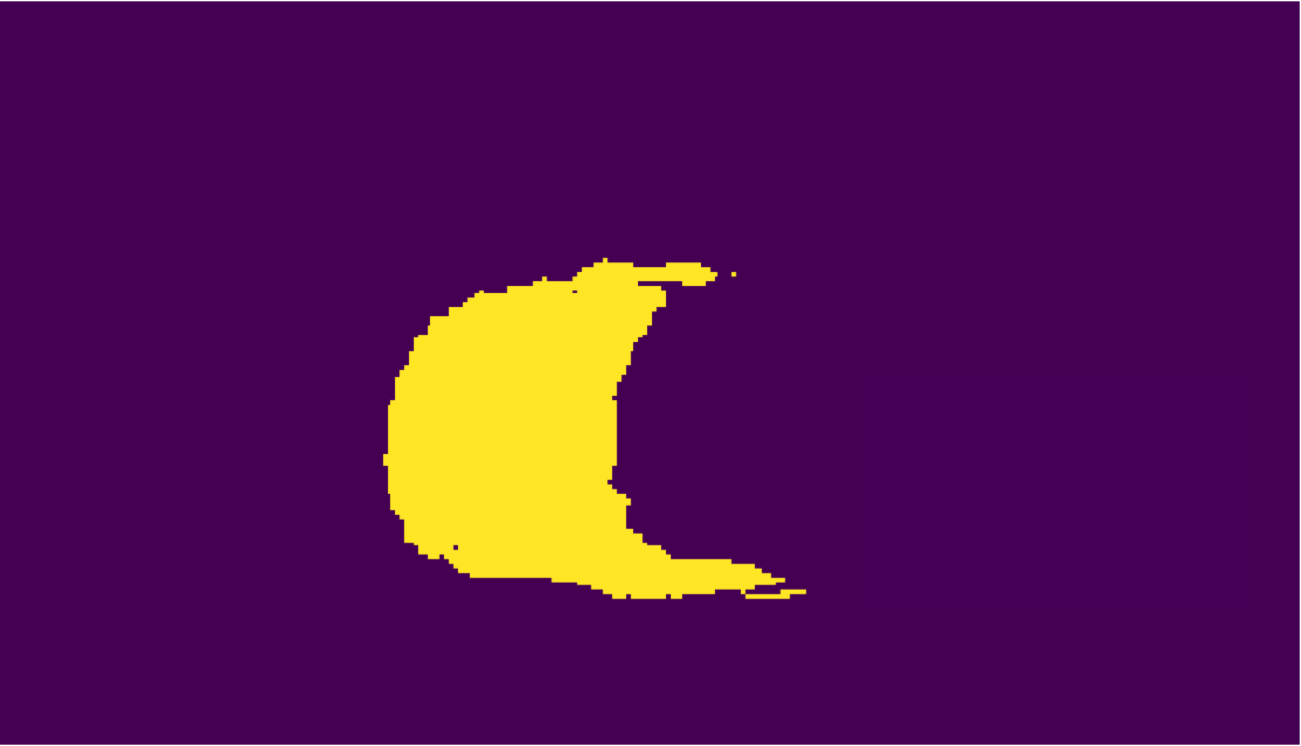}
        \caption{$t=1$, occlusion map.}
        \label{fig:ff4}
    \end{subfigure}%
    \caption{Ghosting effect caused by warping. Consider a foreground object on the background. \textbf{(a)} shows a predicted optical flow $\hat{\vf}_1$ from target frame to current frame. \textbf{(b)} shows ghosting effects on the propagated frame $\tilde{\vx}_1$ on locations to which flow has no projection. \textbf{(c)} shows our pixel density map computed by procedure described by Section \ref{sec:map}. \textbf{(d)}  shows our sparse occlusion map on which occlusion locations are colored as yellow.}
    \label{fig:fail}
\end{figure}

However, $\cF$ introduces ``ghosting'' effect in dis-occluded regions since regions in the target frame to which
extrapolated optical flow has no projection (see Figure \hyperref[fig:teaser]{1d} and Figure \ref{fig:fail} for detailed explanation). 
Intuitively, the ``ghosting'' area should be excluded from the propagated results by an occlusion Map.
It should be noted that similar ideas have been previously explored in \cite{finn2016unsupervised,hao2018controllable}; but in their contexts, occlusion maps are \textit{linearly learned} to compose pixels from different sources. We argue that these regions can be \textit{explicitly computed} based on backward flows, and hence can be directly masked out. We here describe the computation of occlusion map from an energy-based perspective. And we will later demonstrate the effectiveness of our computed map over the learned map in Section \ref{exp:mask}.

The pixel density can be viewed as an energy map: initially, it is uniformly filled for all pixels, and the motion propagation changes its distribution and make the energy map become dense or sparse on different region.

We initialize the energy field of the first frame to be a matrix filled with ones, denoted as $\mathbf{E}^1 = \mathbf{1}^{H\times W}$ for an image of $H \times W$ size. 

Given a flow field, for each pixel in the first frame, the energy unit on each coordinate will be added into its 4 corresponded coordinates in the second frame bilinearly according to the flow field and we then get a new energy field $\mathbf{E}^2$. We consider two special cases for each coordinate $(x, y)$ in the second frame: 
\begin{enumerate}
    \item If $\mathbf{E}^2_{x,y}=0$, there is no pixel moving to this coordinate, which indicates it will be \textit{dis-occluded};
    \item If $\mathbf{E}^2_{x,y} > 2$, there are at least two pixels in the first frame compete for the same location, which suggests it will be \textit{occluded}.
\end{enumerate}

After complete definition of occluded and dis-occluded regions, we can then get the occlusion map according to
\begin{equation}
\label{eq:maskcompute}
    \hat{\vm}_{i,j}=\begin{cases} 1 & 0<\mathbf{E}^2_{i,j}<2, \\
    0 & otherwise.
    \end{cases}
\end{equation}

\subsection{Occlusion Inpainting as Context Encoding} \label{sec:ir}

Given the propagated frame $\tilde{\vx}_t$ and the computed occlusion map $\hat{\vm}_t$, we can now formulate our second modeling stage as context encoding to inpaint the missing pixels that are left out after propagation module. Our inpainting module $\cP$ adopts generally the same network architecture as its propagation counterpart  with partial convolution blocks~\cite{liu2018image}. As illustrated in Figure \hyperref[fig:arch]{3b}, our encoder takes the previous propagated frame $\tilde{\vx}_t$ and its occlusion map $\hat{\vm}_t$ as inputs, producing a latent feature representation. The decoder then takes this feature representation and synthesizes the missing content.

Specifically in the encoder, the partial convolution operators~\cite{liu2018image} mask out invalid pixels and re-normalize features within clean receptive fields only. 
For the extreme cases where all pixels in the receptive field are masked, we will simply return a zero value as the result. 
One important design choice is that the receptive field of the bottleneck should be bigger than the maximal area of the occlusion masks so that the feature map can be free from mask in our bottleneck.

For the decoder, the $k$th layer feature maps are upsampled and linked with the $k$th layer features from encoder counting from the latent layer in reverse order by skip connections. However, this raises a fusion issue that the feature from the encoder is from an image with invalid pixels.  Previously in~\cite{liu2018image}, feature maps and masks are concatenated channel-wisely and handled by new partial convolutions in their decoder. We find this could be improved by directly refilling the occluded encoder features by the upsampled decoder features. This decoding fusion is repeated at each layer from the bottleneck up to our final output $\hat{\vx}_t$.

To train our inpainting module $\cP$, we design all of our losses to be temporal-independent so that the module can focus on the visual quality. In general, our loss terms consists of the following terms.

\begin{enumerate}
    \item The pixel reconstruction loss
{\small
\begin{equation*}
    \cL_{pix}
    = 
    \cL_p(\hat{\vx}_t, \vx_t; \hat{\vm}_t)  
    +
    \beta\cL_p(\hat{\vx}_t, \vx_t; 1 - \hat{\vm}_t),
\end{equation*}}
where $\cL_p$ is defined in Equation \ref{recon_loss}.
    \item The perceptual and style losses in VGG's n-dimensional latent space $\{\vPsi^n\}$ as in \cite{simonyan2014very}.
{\small \begin{equation*}
\begin{aligned}
    \cL_{prc}
    &= 
    \frac1n \sum_n \|  \left[\vpsi(\tilde{\vx}) - \vpsi(\vx)\right] \odot \hat{\vm}_t \|_1 \\
    &+
    \beta \| \left[\vpsi(\tilde{\vx}) - \vpsi(\vx)\right] \odot (1 - \hat{\vm}_t) \|_1, \\
    \cL_{sty}
    &= 
    \frac1n \sum_n \| (\vpsi(\tilde{\vx}) - \vpsi(\vx))^T (\vpsi(\tilde{\vx}) - \vpsi(\vx)) \odot \hat{\vm}_t \|_1 \\
    &+
    \beta \|(\vpsi(\tilde{\vx}) - \vpsi(\vx))^T (\vpsi(\tilde{\vx}) - \vpsi(\vx)) \odot (1 - \hat{\vm}_t) \|_1.
\end{aligned}
\end{equation*}}
    \item The total-variance loss to encourage similar texture in occlusion boundaries
{\footnotesize \begin{equation*}
\begin{aligned}
    \cL_{var}
    &= \sum_{i,j}\sqrt{\Vert \vx_{t(i,j+1)}-\vx_{t(i,j)} \Vert^2_2+ \Vert \vx_{t(i+1,j)}-\vx_{t(i,j)} \Vert^2_2}.
\end{aligned}
\end{equation*}}
    \item The extra semantic loss to enforce layout consistency between segmentation masks, distilled by a pretrained segmentation network~\cite{rotabulo2017place} ~$\vPhi: \cX \rightarrow \cY$, since unconditional image inpainting tends to remove the foreground objects,
{\small \begin{equation*}
\begin{aligned}
    \cL_{seg}
    &= 
    CE(\vPhi(\hat{\vx}_t) \odot \hat{\vm}_t, \vy_t \odot \hat{\vm}_t) \\
    &+ 
    \beta CE\big(\vPhi(\hat{\vx}_t) \odot (1-\hat{\vm}_t), \vy_t \odot (1-\hat{\vm}_t)\big).
\end{aligned}
\end{equation*}}
\end{enumerate}
The total loss for the inpainting module is
{\begin{equation}
\begin{aligned}
    \cL_{\cP} = &\cL_{pix} + \lambda_{prc}\cL_{prc} + \lambda_{sty}\cL_{sty} \\& + \lambda_{var}\cL_{var}
    + \lambda_{seg} \cL_{seg}.
\end{aligned}
\end{equation}}

In the above losses, $\beta$ is our attentive weight for masked regions and all $\lambda$s are fixed hyper-parameter during training. We use a coarse grid search to set $\lambda_{prc} = 0.05, \lambda_{sty}=120, \lambda_{var} = 0.1, \text{and } \lambda_{seg} = 5$ and we find best experimental performance (see Table \ref{tab:beta}) when $\beta=10$.

\subsection{Training} \label{sec:train}
The overall training objective is formulated as

\begin{equation}
\begin{aligned}
    \min_{\cF, \cP} \big( \cL_{\cF} + \cL_{\cP} \big).
\end{aligned}
\end{equation}

Because flow estimation and prediction are hard to learn and sensitive to data biases, we first train our motion propagation module $\cF$ and inpainting module $\cP$ separately. After gradients become stable, we connect the two components together and fine-tune the whole network in an end-to-end fashion.




\section{Experiments} \label{sec:expm}

In this section, we will first introduce our benchmark datasets and baselines from previous works. Then we will verify our model's performance on next-frame predictions and multi-frame predictions. To better understanding how we disentangle propagation and  generation through an occlusion inpainting way, we also investigate the properties of occlusion maps.

\begin{figure*}[t]
\setlength\tabcolsep{2pt}%
\begin{tabularx}{\textwidth}{@{}c*{6}{C}@{}}
   &
   \includegraphics[ width=\linewidth, height=\linewidth, keepaspectratio]{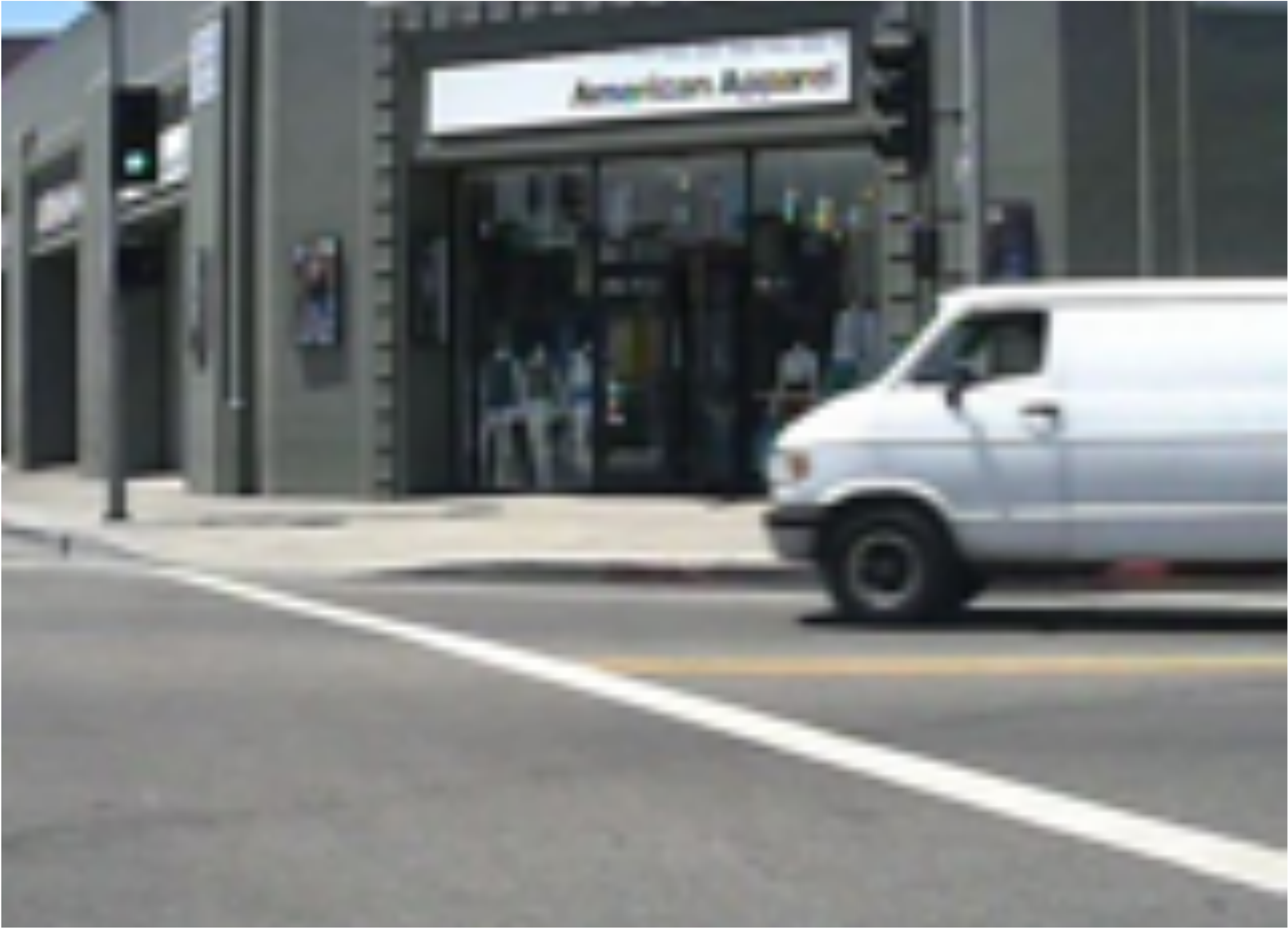} &
   \includegraphics[ width=\linewidth, height=\linewidth, keepaspectratio]{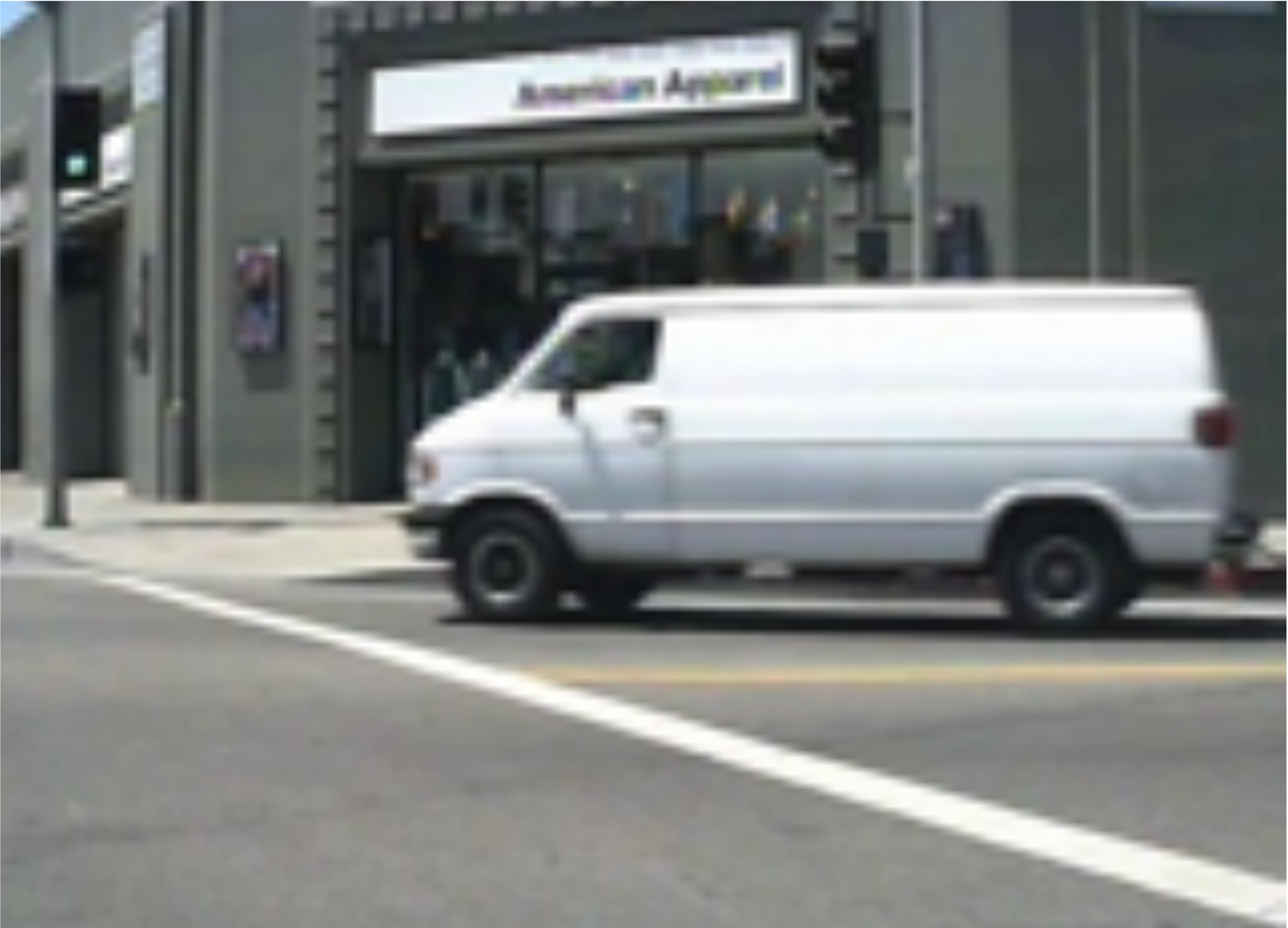} &
   \includegraphics[ width=\linewidth, height=\linewidth, keepaspectratio]{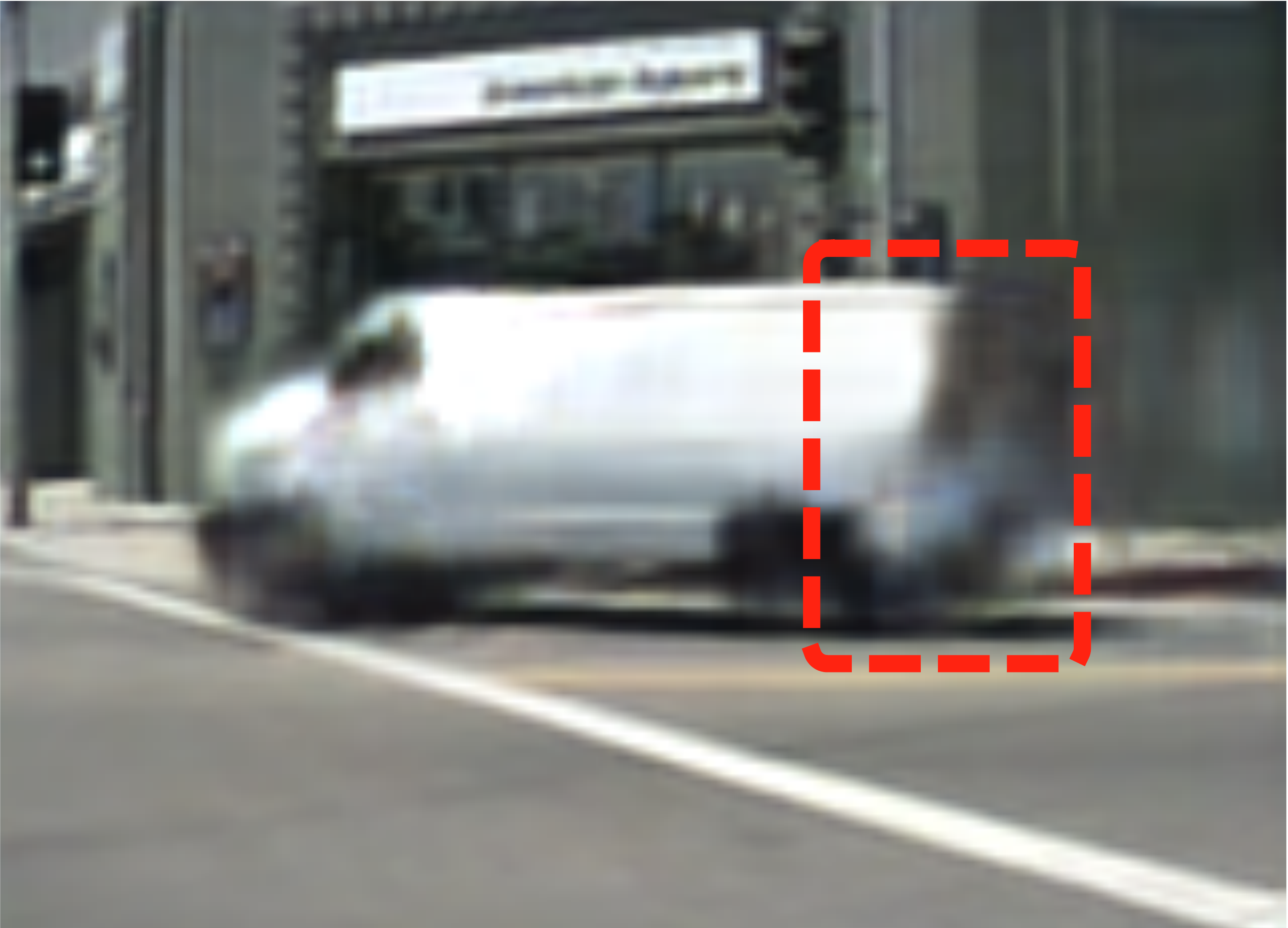} &
   \includegraphics[ width=\linewidth, height=\linewidth, keepaspectratio]{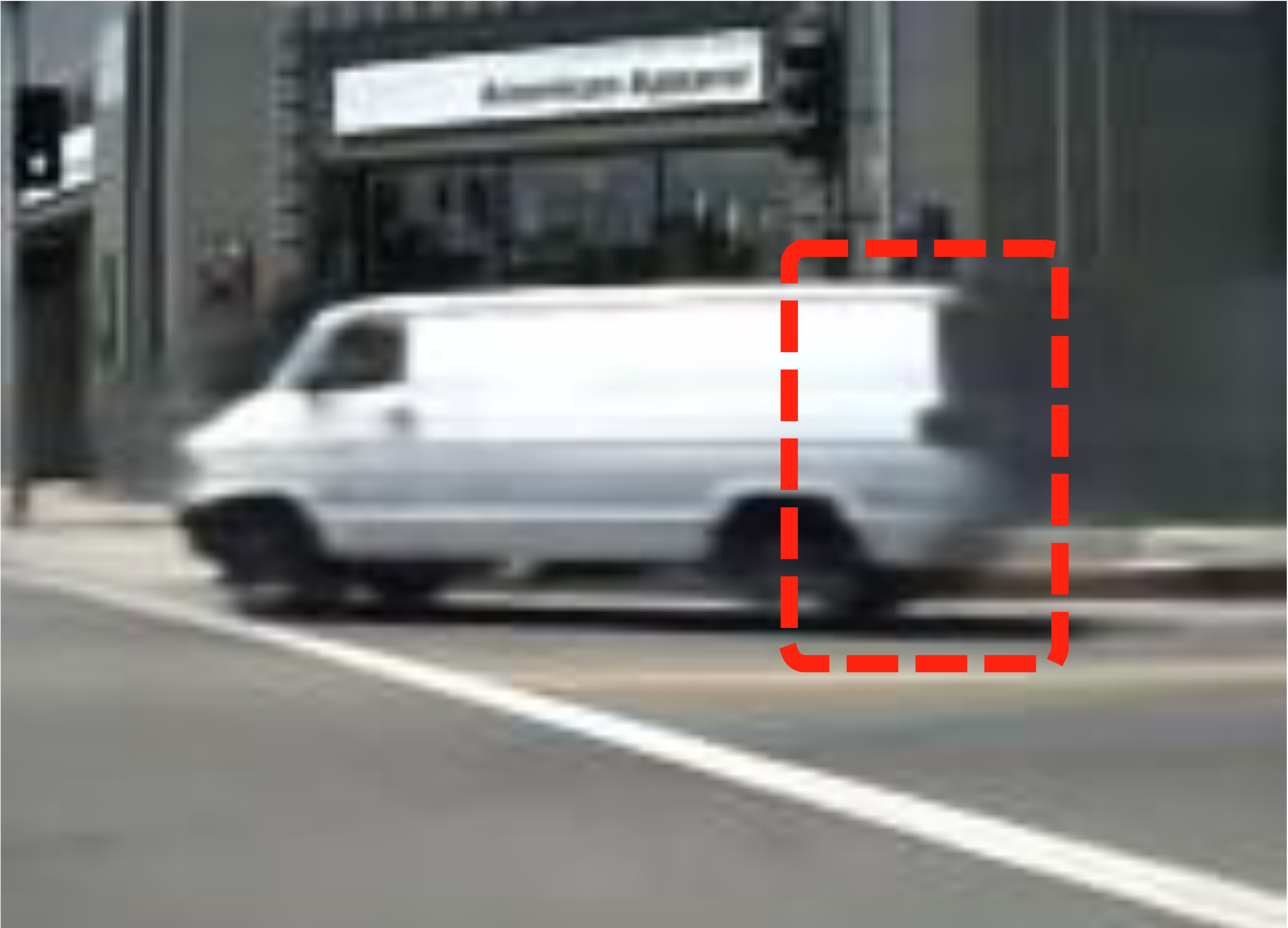} &
   \includegraphics[ width=\linewidth, height=\linewidth, keepaspectratio]{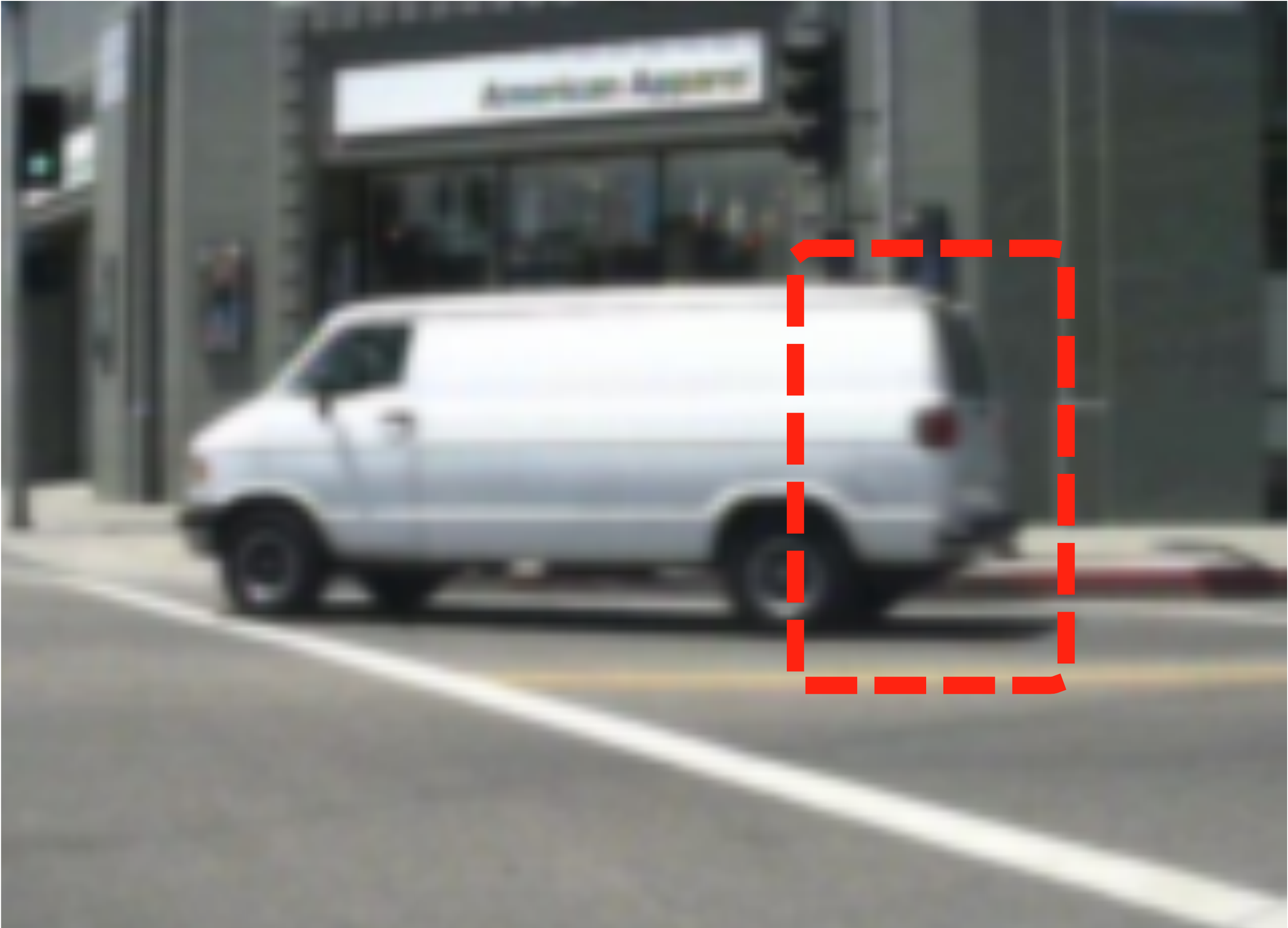} &
   \includegraphics[ width=\linewidth, height=\linewidth, keepaspectratio]{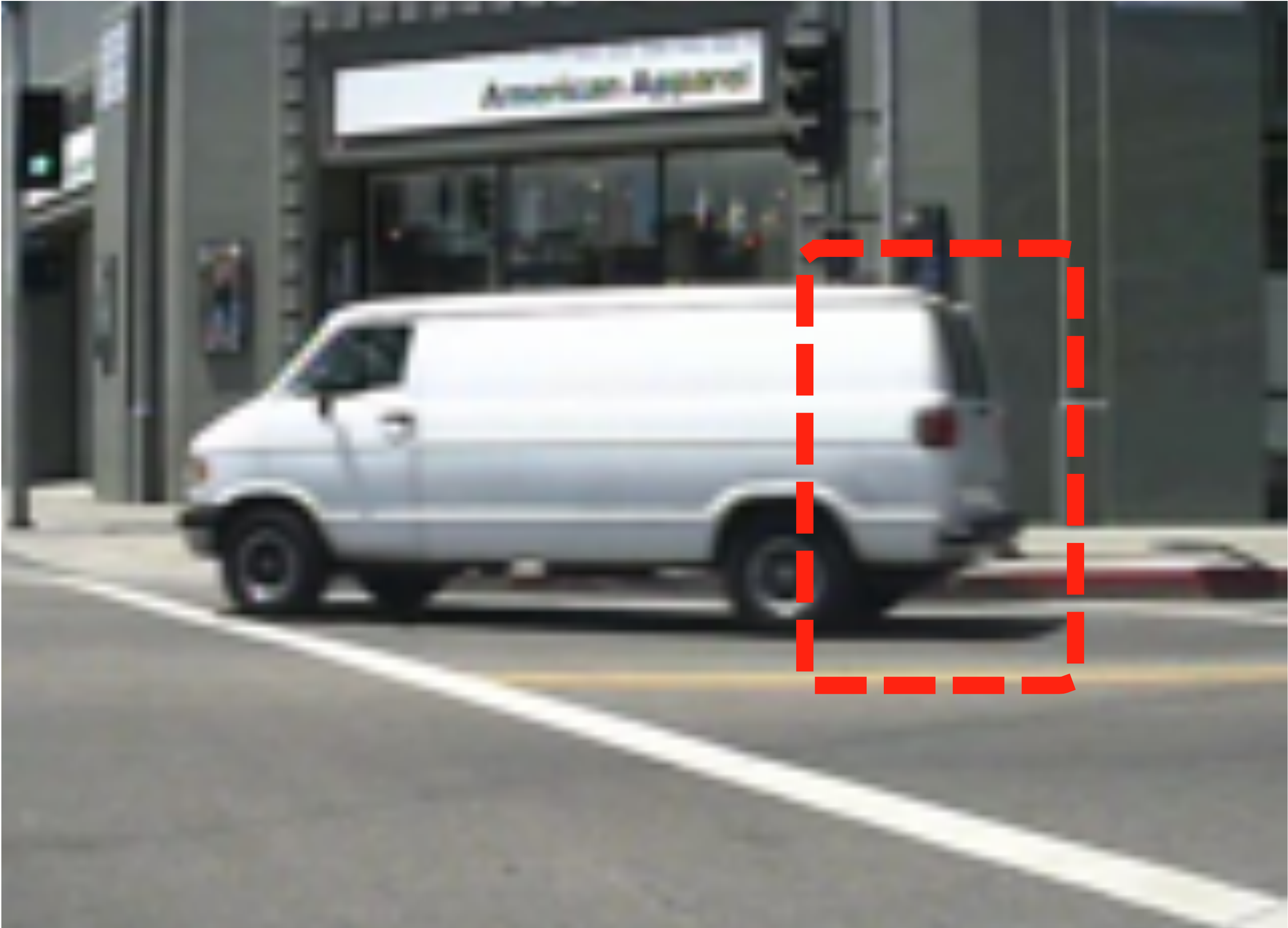} \\
   &
   \includegraphics[ width=\linewidth, height=\linewidth, keepaspectratio]{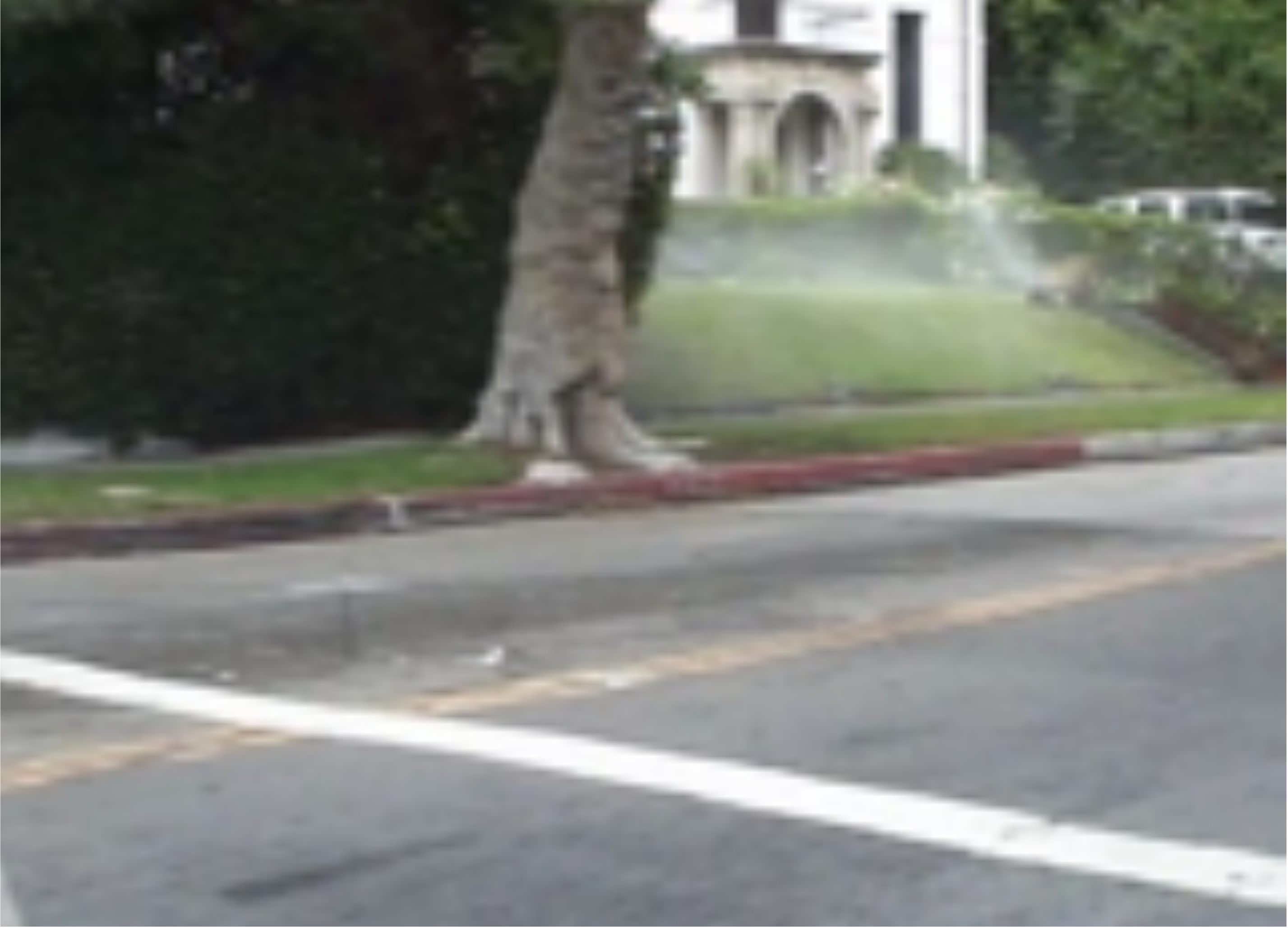} &
   \includegraphics[ width=\linewidth, height=\linewidth, keepaspectratio]{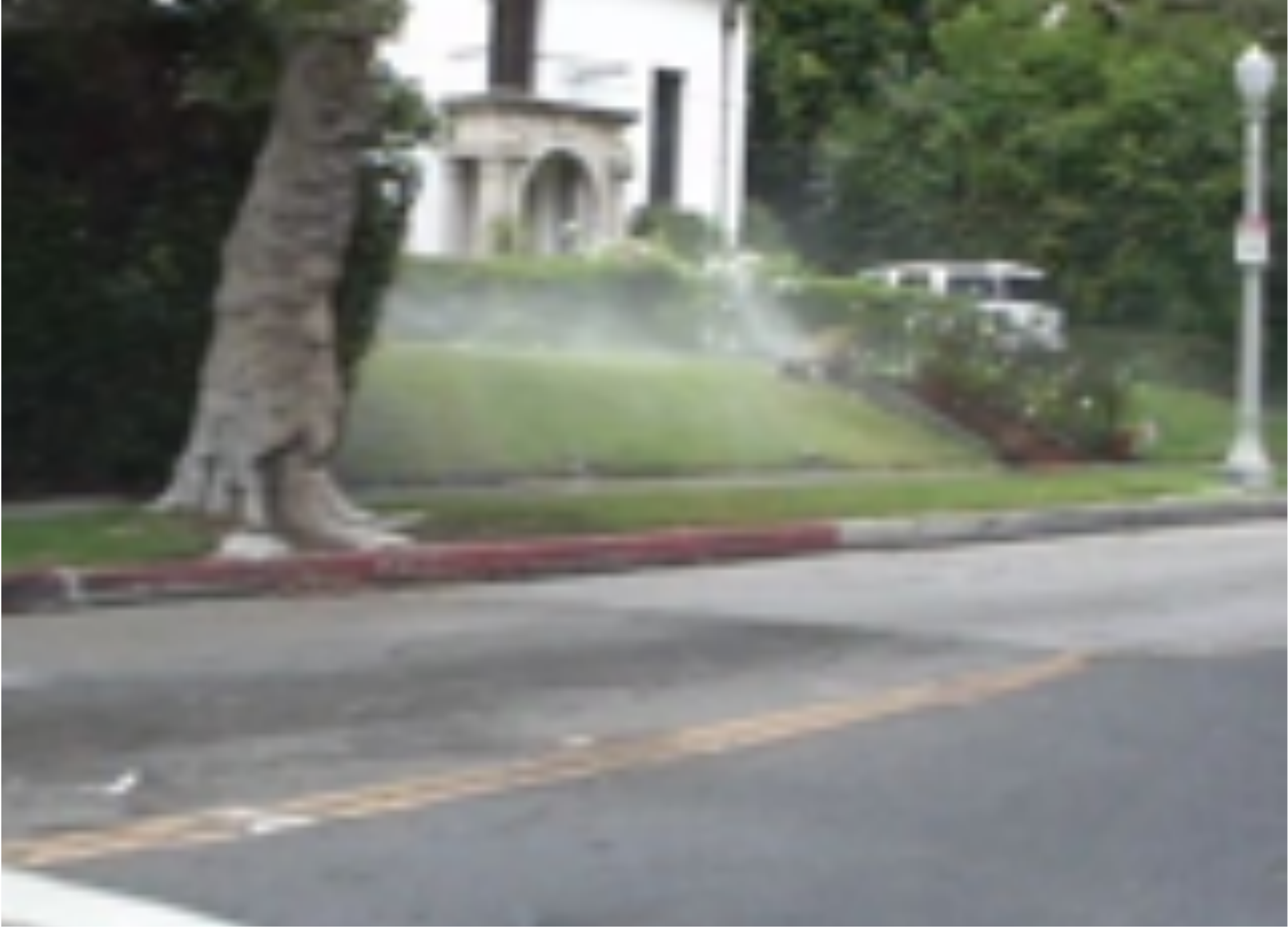} &
   \includegraphics[ width=\linewidth, height=\linewidth, keepaspectratio]{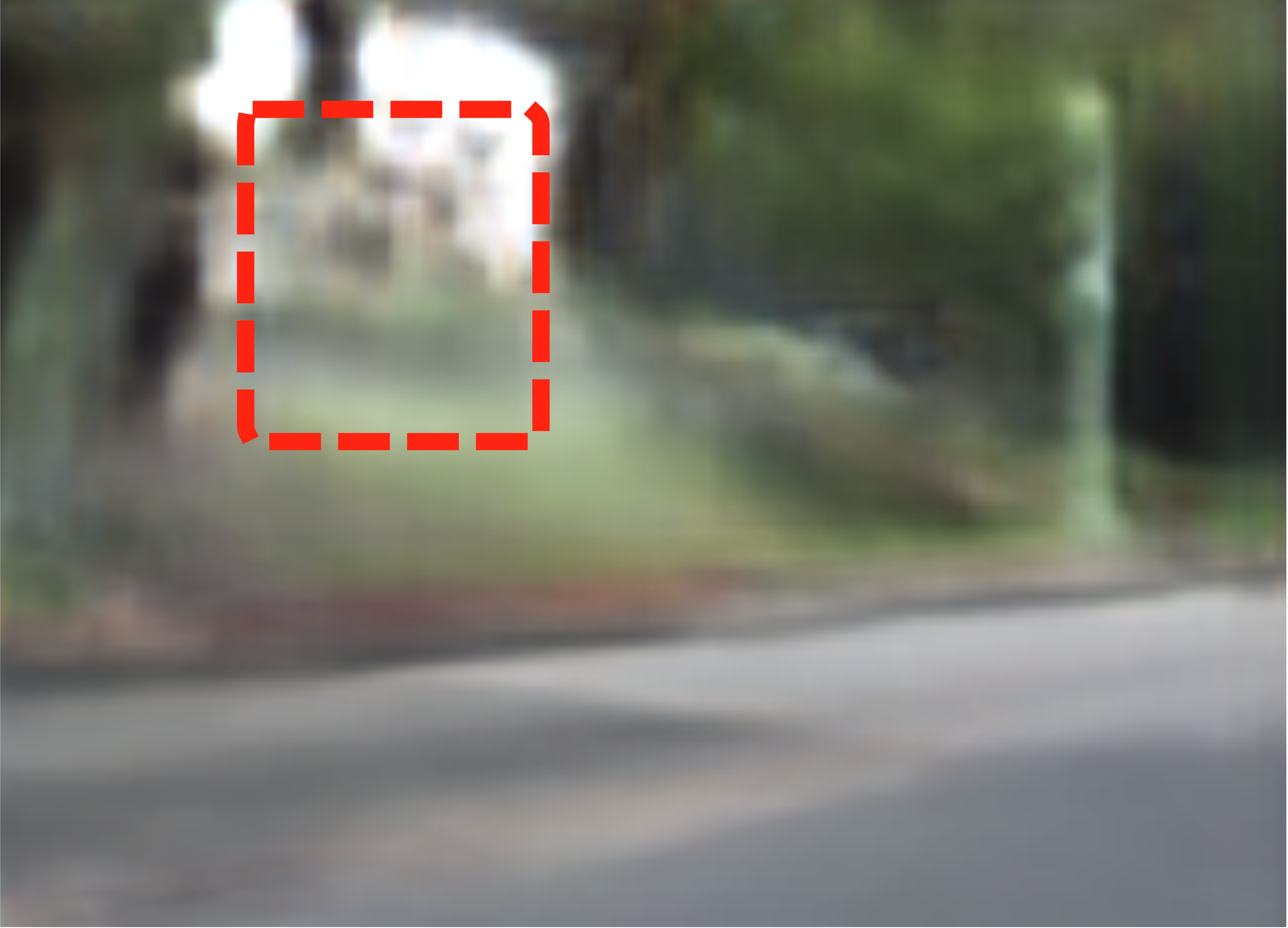} &
   \includegraphics[ width=\linewidth, height=\linewidth, keepaspectratio]{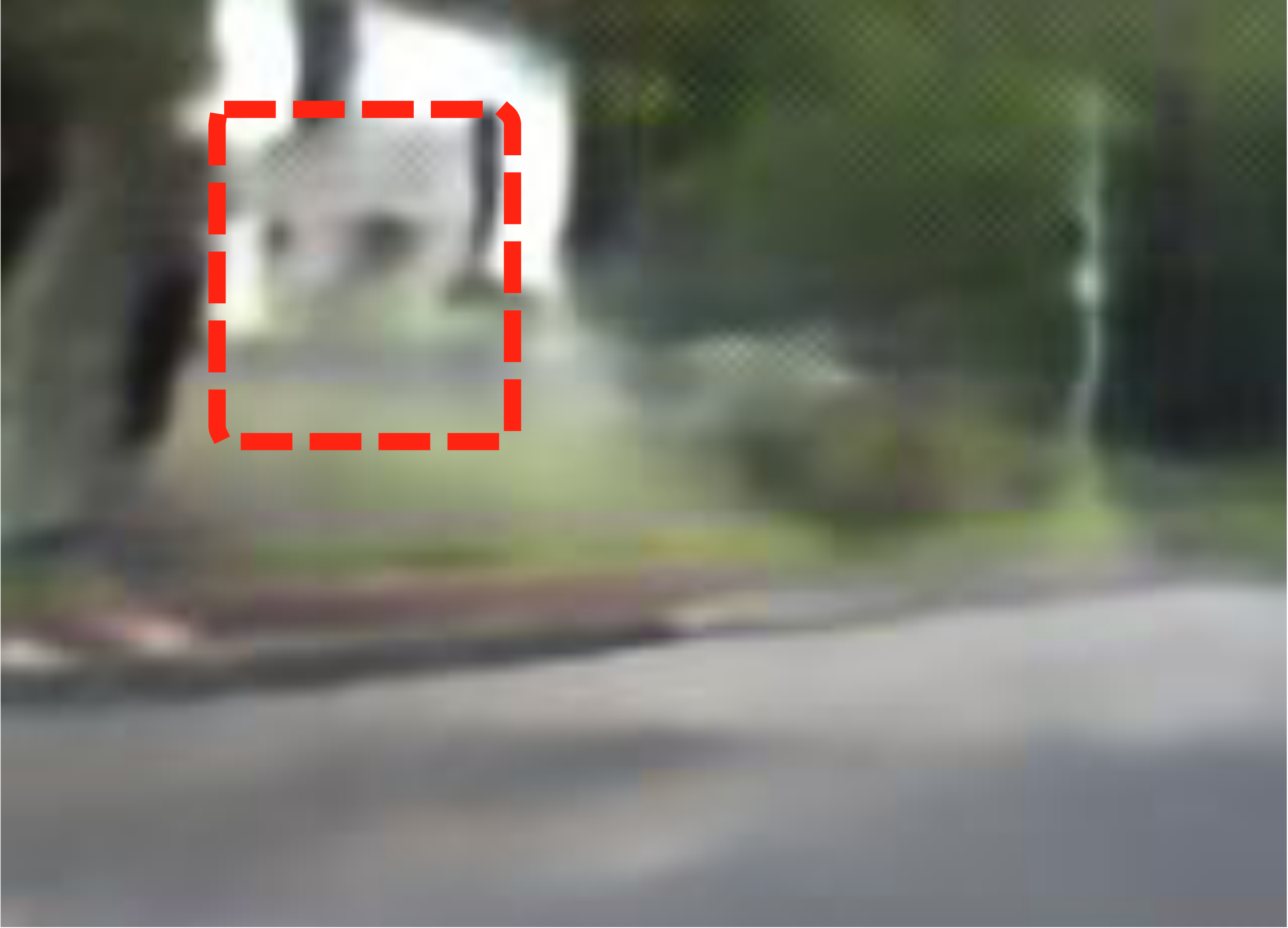} &
   \includegraphics[ width=\linewidth, height=\linewidth, keepaspectratio]{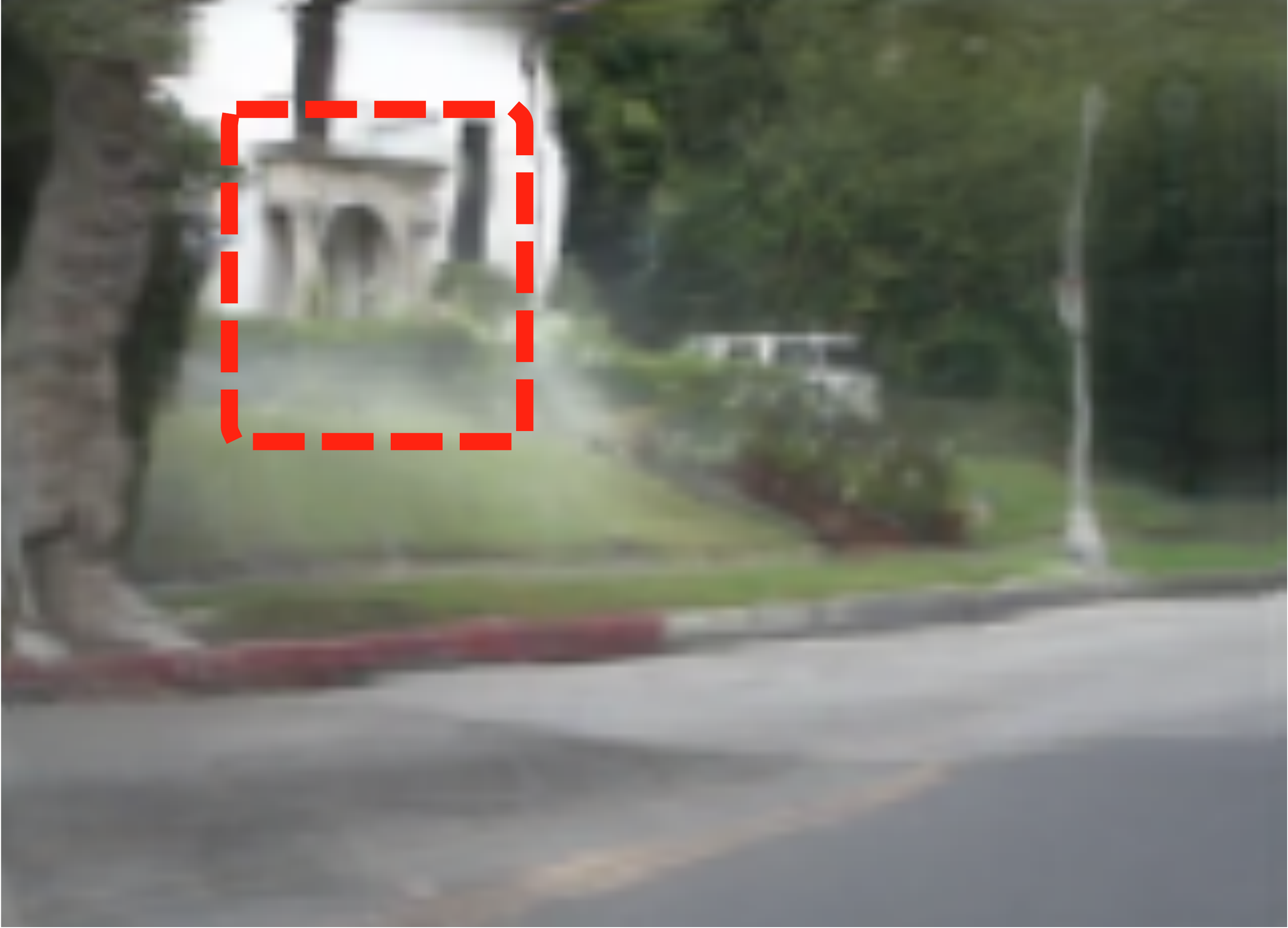} &
   \includegraphics[ width=\linewidth, height=\linewidth, keepaspectratio]{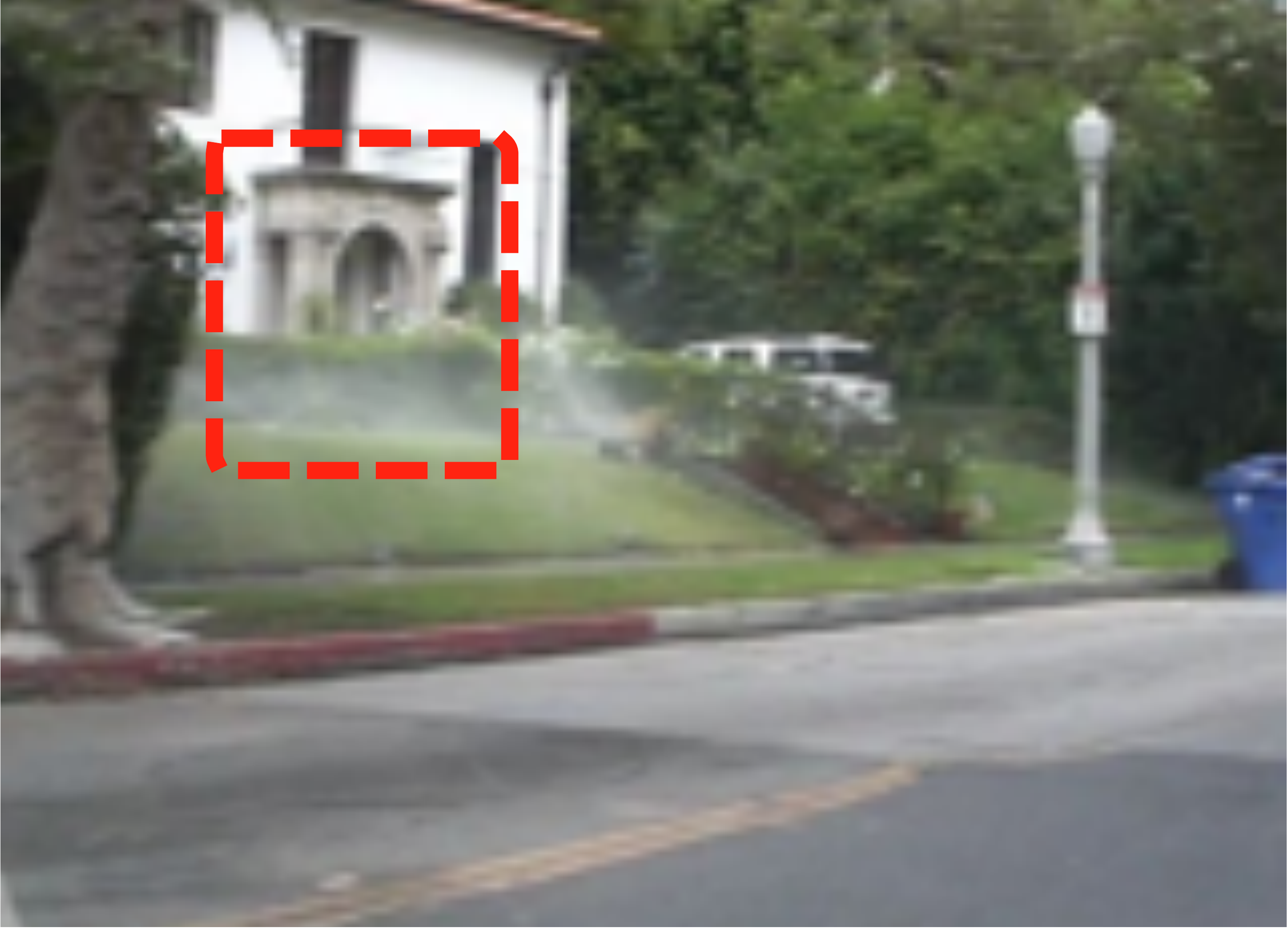} \\
   &
   (a)\enspace $\vx_{t-5}$ &
   (b)\enspace $\vx_{t-2}$ &
   (c)\enspace PredNet &
   (d)\enspace ContextVP &
   (e)\enspace \model{} (Ours) &
   (f)\enspace $\vx_{t}$ \\
\end{tabularx}
    \caption{Qualitative comparisons for Next-Frame Prediction on CalTech Pedestrian dataset.
    Given $10$ past frames, all models are required to predict the future frame in the next step. Our \model{} produces much sharper results compared to previous state-of-the-art methods on this dataset. Remarkably, it is also capable of inferring semantically sensible appearance for occluded regions. Results are best viewed in color with zoom.
    }
    \label{fig:calped-nfp}
\end{figure*}

\subsection{Experiment Setup}
\bfsection{Metrics}
We adopt the existing Peak Signal-to-Noise Ratio (PSNR) and Structural Similarity Index Measure (SSIM)~\cite{wang2004image} metrics to measure the pixel/patch-wise accuracy. However, it is well-known that these metrics disagree with the human perception \cite{wichers2018hierarchical,srivastava2015unsupervised,kalchbrenner2016video,walker2017pose} because they tend to encourage blurriness over naturalness. Therefore we also measure the realism of results by Learned Perceptual Image Patch Similarity (LPIPS\footnote{We use LPIPS Version 0.1 available online.}) proposed by Zhang~\etal~\cite{zhang2018unreasonable}. Higher PSNR/SSIM scores and lower LPIPS distances suggest better performance.

\bfsection{Dataset}
We evaluate our results on the CalTech Pedestrian dataset~\cite{dollar2012pedestrian} and the KITTI Flow dataset~\cite{menze2015object}. The previous setup on the CalTech Pedestrian dataset is to first train a model on the training split of KITTI Raw~\cite{menze2015joint} proposed by Lotter \etal~\cite{lotter2016deep} and then directly test it on the testing set of CalTech Pedestrian. Frames are center-cropped and down-sampled to $128\times160$ pixels. Every $11$ consecutive frames are divided and sampled as a training clip in which the first $10$ frames are fed into the model as the input, and the $11$th frame is used as prediction target. As the results, the training, validation and testing sets consists of $3738$, $14$, and $1948$ clips.

The KITTI Flow dataset is originally designed as a benchmark for optical-flow evaluation. For it is featured with higher resolution, larger motion and more occlusions, we should mention that it is more challenging compared to the raw KITTI dataset. It contains $3823$ examples for training, $378$ for validation and $4167$ for testing. For all input images, we down-sampled and then center-cropped them into $320\times640$ pixels. We apply data augmentation techniques such as random cropping and random horizontal flipping for all the models. In addition, we sample video clips of $5$ frames ($4$-in $1$-out) from the dataset using a sliding window. This amounts to $3500$ clips for training and $4000$ clips for testing.


\bfsection{Baselines}
We consider representative baselines from three model families for video prediction: (1) \textit{pixel-based} methods, including Beyond MSE~\cite{mathieu2015deep}, PredNet~\cite{lotter2016deep}, SVP-LP~\cite{denton2018stochastic}, MCNet~\cite{villegas2017decomposing}, MoCoGAN~\cite{tulyakov2017mocogan}, and ContextVP~\cite{byeon2017contextvp}; (2) \textit{motion-based} method, including DVF~\cite{liu2017video}; (3) \textit{linear fusion} methods, including Dual Motion GAN~\cite{liang2017dual} and CtrlGen~\cite{hao2018controllable}. Note that we have not included \cite{finn2016unsupervised} in our experiments since on its proposed RobotPush datasets, PredNet and SVP-LP are stronger baselines.
For SVP-LP, we sampled 100 trajectories for each clip as in the original paper. The reported results are the mean performance over all the prediction results.

\subsection{Next-Frame Prediction} \label{sec:nfp}
\begin{table}[t]
    \centering
    \resizebox{\linewidth}{!}{%
    \begin{tabular}{ccccc}
        \toprule
        \textbf{Method}
        & \textbf{Family}
        & \textbf{PSNR}$\uparrow$
        & \textbf{SSIM}$\uparrow$
        & \textbf{LPIPS}$\downarrow$ $\mathbf{(\times 10^{-2})}$
        \\
        \midrule
        Repeat 
        & $-$
        & $23.3$
        & $0.779$
        & $5.11$ 
        \\
        \midrule
        BeyondMSE~\cite{mathieu2015deep}
        & \textit{P}
        & $-$   
        & $0.881$ 
        & $-$           
        \\
        PredNet~\cite{lotter2016deep} 
        & \textit{P}
        & $27.6$ 
        & $0.905$ 
        & $7.47$   
        \\
        ContextVP~\cite{byeon2017contextvp}  
        & \textit{P}
        & $\mathbf{28.7}$ 
        & $0.921$ 
        & $6.03$   
        \\
        \midrule
        DVF~\cite{liu2017video}    
        & \textit{M}
        & $26.2$ 
        & $0.897$ 
        & $5.57$    
        \\
        \midrule
        Dual Motion GAN~\cite{liang2017dual} 
        & \textit{F}
        & $-$  
        & $0.899$ 
        & $-$      
        \\
        CtrlGen~\cite{hao2018controllable}  
        & \textit{F}
        & $26.5$  
        & $0.900$ 
        & $6.38$      
        \\
        \model{} (Ours)
        & \textit{F}
        & $28.2$ 
        & $\mathbf{0.923}$
        & $\mathbf{5.04}$ 
        \\
        \bottomrule
    \end{tabular}
    }
    \caption{
    Next-Frame Prediction results on CalTech Pedestrian. All models are trained on KITTI Raw dataset. The best results under each metric are marked in bold. 
    }
    \label{tab:calped-nfp}
    \vspace{-0.4cm}
\end{table}
\begin{figure*}[htb!]
\setlength\tabcolsep{2pt}%
\begin{tabularx}{\textwidth}{@{}c*{4}{C}@{}}
   &
   \includegraphics[ width=\linewidth, height=\linewidth, keepaspectratio]{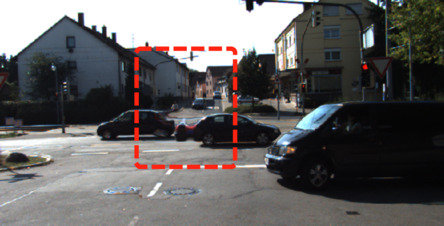} &
   \includegraphics[ width=\linewidth, height=\linewidth, keepaspectratio]{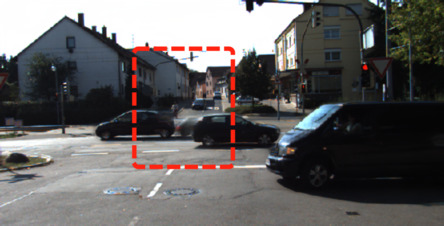} &
   \includegraphics[ width=\linewidth, height=\linewidth, keepaspectratio]{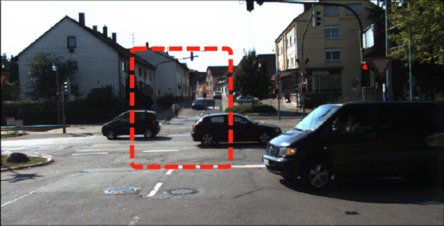} &
   \includegraphics[ width=\linewidth, height=\linewidth, keepaspectratio]{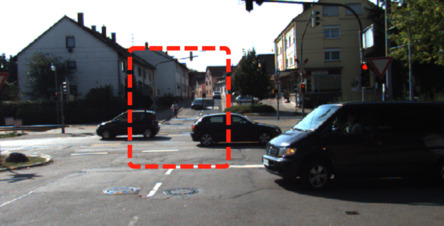} \\
   &
   \includegraphics[ width=\linewidth, height=\linewidth, keepaspectratio]{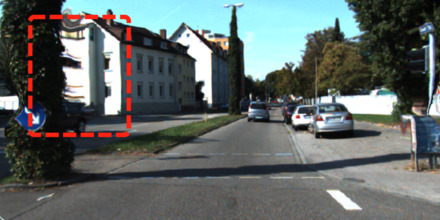} &
   \includegraphics[ width=\linewidth, height=\linewidth, keepaspectratio]{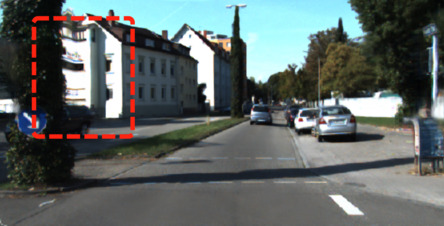} &
   \includegraphics[ width=\linewidth, height=\linewidth, keepaspectratio]{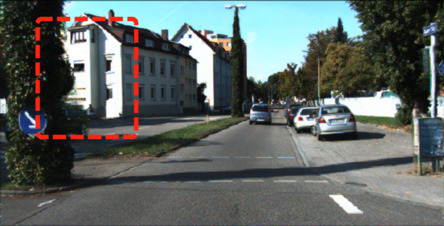} &
   \includegraphics[ width=\linewidth, height=\linewidth, keepaspectratio]{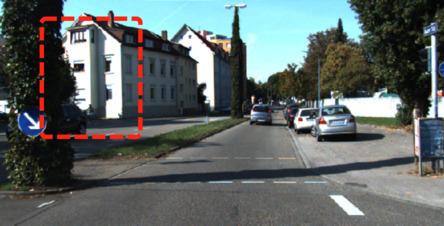} \\
   &
   (a)\enspace DVF & 
   (b)\enspace CtrlGen &
   (c)\enspace \model{} (Ours) &
   (d)\enspace $\vx_t$ \\
\end{tabularx}
    \caption{Qualitative comparisons for Next-Frame Prediction on the more challenging KITTI Flow dataset. All models are given 4 frames as input and required to predict the next frame. Frames include large motions, scene changes or occlusions. Our method is robust to these cases and consistent in the performance. More results please refer to our supplementary.}
    \label{fig:qual-nfp}
\end{figure*}

We begin our evaluations on next-frame prediction. The goal of our task is simple and intuitive: Given the input frames, we need to output the next frame as accurate as possible.

Our result on CalTech Pedestrian dataset is shown on Table \ref{tab:calped-nfp}. We compare our model against previous state-of-the-art methods on this dataset. Our model achieves comparable PSNR and SSIM scores with ContextVP \cite{byeon2017contextvp}. Meanwhile, our method can predict non-stretching textures in those occluded regions, which leads to smaller perceptual dissimilarity measured by LPIPS. As shown in Figure \ref{fig:calped-nfp}, our model is robust for both pixel propagation and novel scene inference.

Apart from empirical improvements, we find that, in terms of LPIPS metric, all the evaluated state-of-the-art methods do no better than the most naive baseline --- repeating the last input frame as the prediction. This suggests that the CalTech Pedestrian dataset consists of small motions that are not obvious for human perception. This motivates us to work on KITTI Flow dataset, which is more challenging so that learners can benefit from more inductive biases and thus be more robust.

Table \ref{tab:kflow-nfp} shows our results on KITTI Flow dataset. 
As resolution increases, previous pixel-based methods (PredNet, SVP-LP) suffer from a steeper learning curve and more uncertainty in the visual space, resulting in the noticeable drop in their performance. 
Though achieving the better pixel/patch accuracy, they underperform the weakest repeating baseline in terms of the perceptual similarity. 
Our \model{} achieves the best results in all metrics, especially LPIPS, showing around $30\%$ improvement over the second best result from DVF.

As demonstrated in Figure~\ref{fig:qual-nfp}, our proposed model again produce more visually appealing predictions than our baselines. In contrast to the pixel-based methods, all demonstrated methods suffer less from blurriness but display the distortion and stretch in shapes due to quick scene changes, which cause inaccurate flow prediction. Our model, instead, can predict better flow so as to alleviate undesirable artifacts in large motion areas. Occluded areas are masked by motion propagation and refilled by in-painting so that they are free of ghosting effects. Our occlusion in-painter learns a scene prior to hallucinate what is missing given the contextual information.

It is interesting to see that our model can achieve better PSNR/SSIM scores with training emphasized on realism terms (perceptual, style, and segmentation).
We argue that this could serve as the evidence that our method could effectively learn dataset priors and flow prediction given the same amount of data.


\begin{table}[t] 
    \centering
    \resizebox{\linewidth}{!}{%
    \begin{tabular}{ccccc}
        \toprule
        \textbf{Method}
        & \textbf{Family}
        & \textbf{PSNR}$\uparrow$
        & \textbf{SSIM}$\uparrow$
        & \textbf{LPIPS}$\downarrow$ $\mathbf{(\times 10^{-2})}$
        \\
        \midrule
        Repeat 
        & $-$
        & $16.5$
        & $0.489$
        & $19.0$ 
        \\
        \midrule
        PredNet \cite{lotter2016deep} 
        & \textit{P}
        & $17.0$ 
        & $0.527$ 
        & $26.3$   
        \\
        SVP-LP \cite{denton2018stochastic} 
        & \textit{P}
        & $18.5$  
        & $0.564$ 
        & $20.2$      
        \\
        MCNet \cite{villegas2017decomposing}
        & \textit{P}
        & $18.9$  
        & $0.587$ 
        & $23.7$      
        \\
        MoCoGAN \cite{tulyakov2017mocogan}
        & \textit{P}
        & $19.2$  
        & $0.572$ 
        & $18.6$      
        \\
        \midrule
        DVF \cite{liu2017video}
        & \textit{M}
        & $22.1$   
        & $0.683$ 
        & $16.3$           
        \\
        \midrule
        CtrlGen \cite{hao2018controllable}    
        & \textit{F}
        & $21.8$ 
        & $0.678$ 
        & $17.9$    
        \\
        \model ~(Ours)
        & \textit{F}
        & $\mathbf{22.3}$ 
        & $\mathbf{0.696}$ 
        & $\mathbf{11.4}$ 
        \\
        \bottomrule
    \end{tabular}
    }
    \caption{
    Next-Frame Prediction results on KITTI Flow. All models are trained to predict next frame given a history buffer of two frames. All evaluation results of the previous methods are obtained by their published codebases. The best results under each metric are marked in bold.
    }
    \label{tab:kflow-nfp}
    \vspace{-0.4cm}
\end{table}
\begin{figure}[tp]
    \centering
    \begin{subfigure}[b]{0.43\linewidth}
        \includegraphics[width=\textwidth]{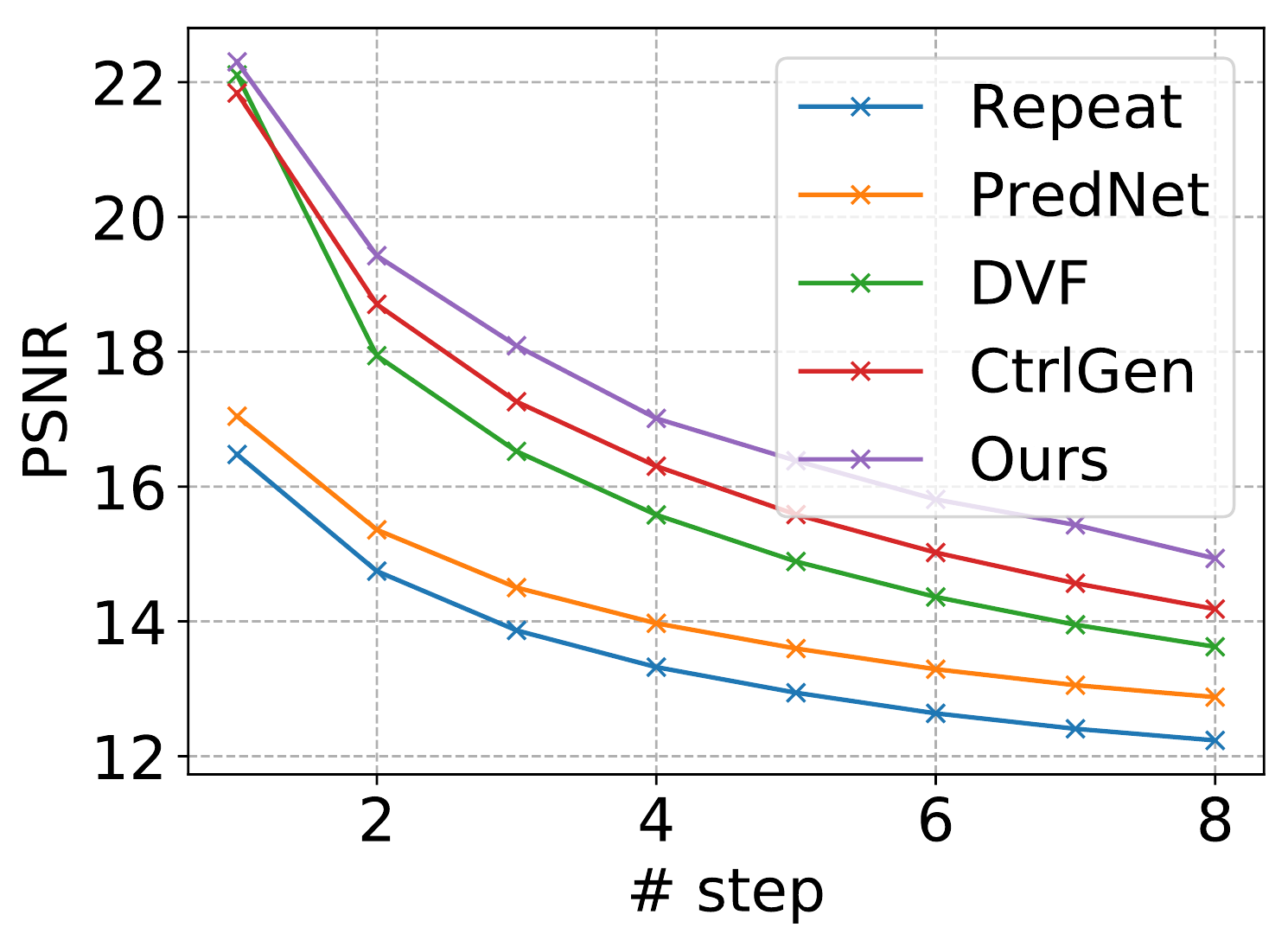}
        \caption{PSNR$\uparrow$}
    \end{subfigure} 
    \begin{subfigure}[b]{0.43\linewidth}
        \includegraphics[width=\textwidth]{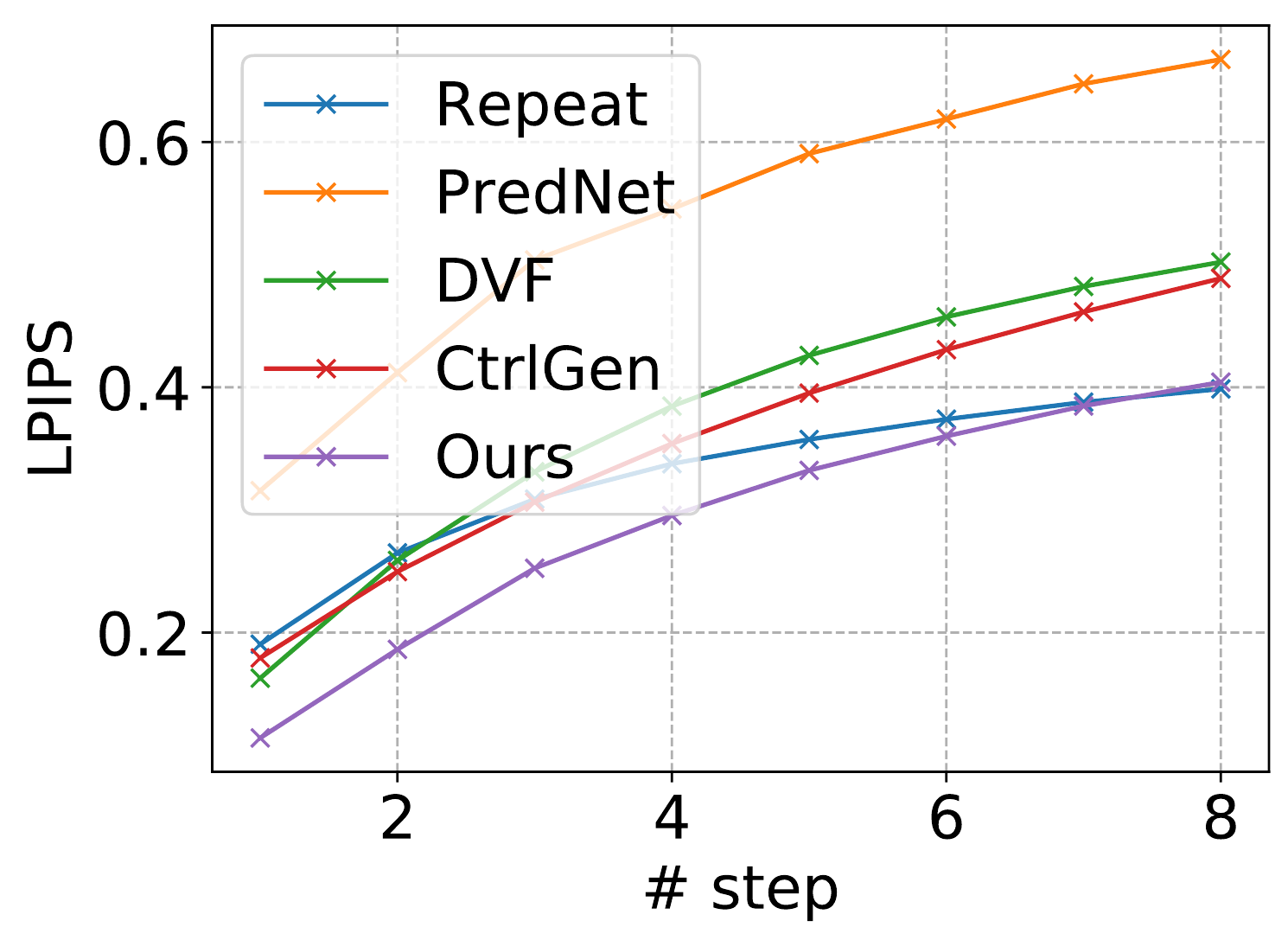}
        \caption{LPIPS$\downarrow$}
    \end{subfigure}
    \caption{Quantitative results for Multi-Frame Prediction on KITTI Flow dataset.
    All the models take in $4$ frames as input and recursively predict next $8$ frames. Best viewed in color with zoom. For qualitative results and SSIM quantitative result, please refer to our supplementary materials.}
    \label{fig:quant-mfp}
\end{figure}
\subsection{Multi-Frame Prediction} \label{sec:kflow}
We next move to the more challenging Multi-frame Prediction task. Comparing to Next-Frame Prediction, it requires our models to output a sequence of frames to match the ground-truth future frames. To do a fair comparison between our method and other baselines, we limit all input frames to be 4 frames and requires each model to predict 8 following frames.

Figure~\ref{fig:quant-mfp} compares the results on Multi-Frame Prediction of our models with various baselines on KITTI Flow dataset. Our method shows consistent performance gains on all metrics through time. DVF performs similarly to our model for short-term prediction measured by PSNR but quickly decays after $2$ steps. This is because that their method is sensitive to propagated error since there are no remedy mechanisms. Our model, however, can mask out undesirable regions and in-paint new pixels instead.


\subsection{Ablations on Occlusion Map} \label{sec:disc}

\label{exp:mask}
To better evaluate the effectiveness of our computed occlusion map, we finally design a progressive curriculum for ablations. 

\noindent \textbf{Does occlusion map really help?} \quad We design different
baselines to verify our design choices for computing the occlusion map. Results
are shown on Table \ref{tab:mask1}. The first baseline directly warps the last
input frame as in \cite{liu2017video}, which also produces the result of our
motion propagation module. Next baseline deploys an Auto-Encoder to directly
refine warped frames without reasoning about occlusion map. The third one learns
an independent occlusion map so as to fuse the result of motion propagation
module and occlusion in-painter module, similar to \cite{hao2018controllable},
in an unsupervised way. It is interesting that the performance will downgrade if
we only refine the motion propagated results without any guidance. Also, we find
that without occlusion supervision in the wild, our computed occlusion map help
final prediction performance better than its learned counterpart.

\begin{table}
    \centering
    \resizebox{\linewidth}{!}{
    \begin{tabular}{cccc}
        \toprule
        \textbf{Method}
        & \textbf{PSNR}$\uparrow$
        & \textbf{SSIM}$\uparrow$
        & \textbf{LPIPS}$\downarrow$
        \\
        \midrule
        Warped only~\cite{liu2017video}
        & $21.6$ 
        & $0.684$ 
        & $0.139$ 
        \\
        No occlusion map (auto-encoder)
        & $21.3$ 
        & $0.674$ 
        & $0.120$
        \\
        Learned occlusion map~\cite{hao2018controllable}
        & $21.8$ 
        & $0.679$ 
        & $0.179$
        \\
        \midrule
        Sparse occlusion map (Ours)
        & $\mathbf{22.3}$ 
        & $\mathbf{0.696}$ 
        & $\mathbf{0.114}$ 
        \\
        \bottomrule
    \end{tabular}
    }
    \caption{
    Ablation study on different design choice for occlusion map.
    }
    \label{tab:mask1}
\end{table}

\noindent \textbf{Why does occlusion map help?} \quad We are interested in how
big is the gap between the their localization qualities. Since it is hard to get
the ground truth occlusion map in the real world. We use a synthetic dataset
instead. Here we use RoamingImages~\cite{Janai2018ECCV} dataset for this
specific quantitative study. The benchmark contains $80,000$ examples as moving
a random foreground image in front of a random background image according to a
random linear motion. We report the IoU between the predicted occlusion map and
the ground truth. Results are shown on Table \ref{tab:iou}. Not surprisingly,
our computed occlusion map has much better quality compared to
\cite{hao2018controllable}. Another striking finding is that by also taking
account of the results we shown in Table \ref{tab:mask1}, the performance of
learned occlusion maps is so bad that it can only marginally improve or even
harm downstream prediction performance. 

\begin{table}[]
\centering
    \resizebox{\linewidth}{!}{
\begin{tabular}{c|cc}
\toprule
\small{\textbf{Method}}
& Learned occlusion map~\cite{hao2018controllable}
& Sparse occlusion map (Ours) \\
IoU$\uparrow$ 
& 0.0411 
& 24.91 \\ \bottomrule
\end{tabular}
}
\caption{
Evaluation for learned occlusion map and our sparse occlusion map against ground-truth occlusion map.}
\label{tab:iou}
\end{table}

\noindent \textbf{How does occlusion map help?} \quad We have shown the
effectiveness of the computed occlusion map comparing to other design choices.
Here we take a closer look to the importance of masking weight $\beta$ in all of
our training losses. As shown on Table \ref{tab:beta}, it is meaningful to use a
larger loss weight for the occluded region during the training procedure. 

\begin{table}
    \centering
    \resizebox{\linewidth}{!}{
    \begin{tabular}{cccc}
        \toprule
        \textbf{Method}
        & \textbf{PSNR}$\uparrow$
        & \textbf{SSIM}$\uparrow$
        & \textbf{LPIPS}$\downarrow$
        \\
        \midrule
        $\beta = 0$ (unmasked pixel only)
        & $20.7$ 
        & $0.663$ 
        & $0.159$ 
        \\
        $\beta = 0.1$
        & $20.9$ 
        & $0.671$ 
        & $0.134$ 
        \\
        $\beta = 1$ (same as unmasked pixel)
        & $21.2$ 
        & $0.675$ 
        & $0.123$ 
        \\
        $\beta = \infty$ (masked pixel only)
        & $18.4$
        & $0.514$ 
        & $0.206$ \\
        \midrule
        $\beta = 10$ (Ours)
        & $\mathbf{22.3}$ 
        & $\mathbf{0.696}$ 
        & $\mathbf{0.114}$ 
        \\
        \bottomrule
    \end{tabular}
    }
    \caption{
    Ablation study on different masking weight $\beta$.
    }
    \label{tab:beta}
    \vspace{-0.4cm}
\end{table}

\section{Conclusion}

In this work, we present a method for video prediction by disentangling
motion-specific propagation and motion-agnostic generation. We propose a
disentangled fusion pipeline which gates two tasks into two separate modules and
directly evaluate pixel density map after warping for sparse occlusion maps.
Experiments on synthetic and real datasets show our effectiveness against prior
works.

{\small
\bibliographystyle{ieeetr}
\bibliography{reference}
}

\end{document}